\documentclass[sigconf]{acmart}

\usepackage{subfigure}
\usepackage{graphicx}
\usepackage{multirow}
\usepackage{balance}

\AtBeginDocument{%
  \providecommand\BibTeX{{%
    \normalfont B\kern-0.5em{\scshape i\kern-0.25em b}\kern-0.8em\TeX}}}

\copyrightyear{2024}
\acmYear{2024}
\setcopyright{rightsretained}
\acmConference[SSDBM 2024]{36th International Conference on Scientific and Statistical Database Management}{July 10--12, 2024}{Rennes, France}
\acmBooktitle{36th International Conference on Scientific and Statistical Database Management (SSDBM 2024), July 10--12, 2024, Rennes, France}
\acmPrice{}
\acmDOI{10.1145/3676288.3676292}
\acmISBN{979-8-4007-1020-9/24/07}

\begin{document}

\title{Imbalanced Graph-Level Anomaly Detection via Counterfactual Augmentation and Feature Learning}

\author{Zitong Wang}
\authornote{Both authors contributed equally to this research.}
\affiliation{%
  \institution{School of Cyber Science and Engineering, Wuhan University}
  \city{Wuhan}
  \state{Hubei}
  \country{China}}
\email{zitongwang@whu.edu.cn}

\author{Xuexiong Luo}
\authornotemark[1]
\affiliation{%
  \institution{School of Computing, \\Macquarie University}
  \city{Sydney}
  \state{NSW}
  \country{Australia}}
\email{xuexiong.luo@hdr.mq.edu.au}

\author{Enfeng Song}
\affiliation{%
  \institution{Renmin Hospital of \\Wuhan University}
  \city{Wuhan}
  \state{Hubei}
  \country{China}}
\email{songef@126.com}

\author{Qiuqing Bai}
\affiliation{%
  \institution{School of Cyber Science and Engineering, Wuhan University}
  \city{Wuhan}
  \state{Hubei}
  \country{China}}
\email{baiqiuqing@whu.edu.cn}

\author{Fu Lin}
\authornote{Corresponding author.}
\affiliation{
  \institution{School of Computer Science,\\School of Cyber Science and Engineering, Wuhan University}
  \city{Wuhan}
  \state{Hubei}
  \country{China}}
\email{linfu@whu.edu.cn}

\renewcommand{\shortauthors}{Wang and Luo, et al.}

\begin{abstract}
  Graph-level anomaly detection (GLAD) has already gained significant importance and has become a popular field of study, attracting considerable attention across numerous downstream works. The core focus of this domain is to capture and highlight the anomalous information within given graph datasets. In most existing studies, anomalies are often the instances of few. The stark imbalance misleads current GLAD methods to focus on learning the patterns of normal graphs more, further impacting anomaly detection performance. Moreover, existing methods predominantly utilize the inherent features of nodes to identify anomalous graph patterns which is approved suboptimal according to our experiments. In this work, we propose an imbalanced GLAD method via counterfactual augmentation and feature learning. Specifically, we first construct anomalous samples based on counterfactual learning, aiming to expand and balance the datasets. Additionally, we construct a module based on Graph Neural Networks (GNNs), which allows us to utilize degree attributes to complement the inherent attribute features of nodes. Then, we design an adaptive weight learning module to integrate features tailored to different datasets effectively to avoid indiscriminately treating all features as equivalent. Furthermore, extensive baseline experiments conducted on public datasets substantiate the robustness and effectiveness. Besides, we apply the model to brain disease datasets, which can prove the generalization capability of our work. The source code of our work is available online\footnote{The source code is at \url{https://github.com/whb605/IGAD-CF.git}}.
\end{abstract}



\maketitle

\section{Introduction}
\begin{figure*}[!h]
\centering
\subfigure[Distributions on the normal samples]{
\includegraphics[width=0.3\textwidth]{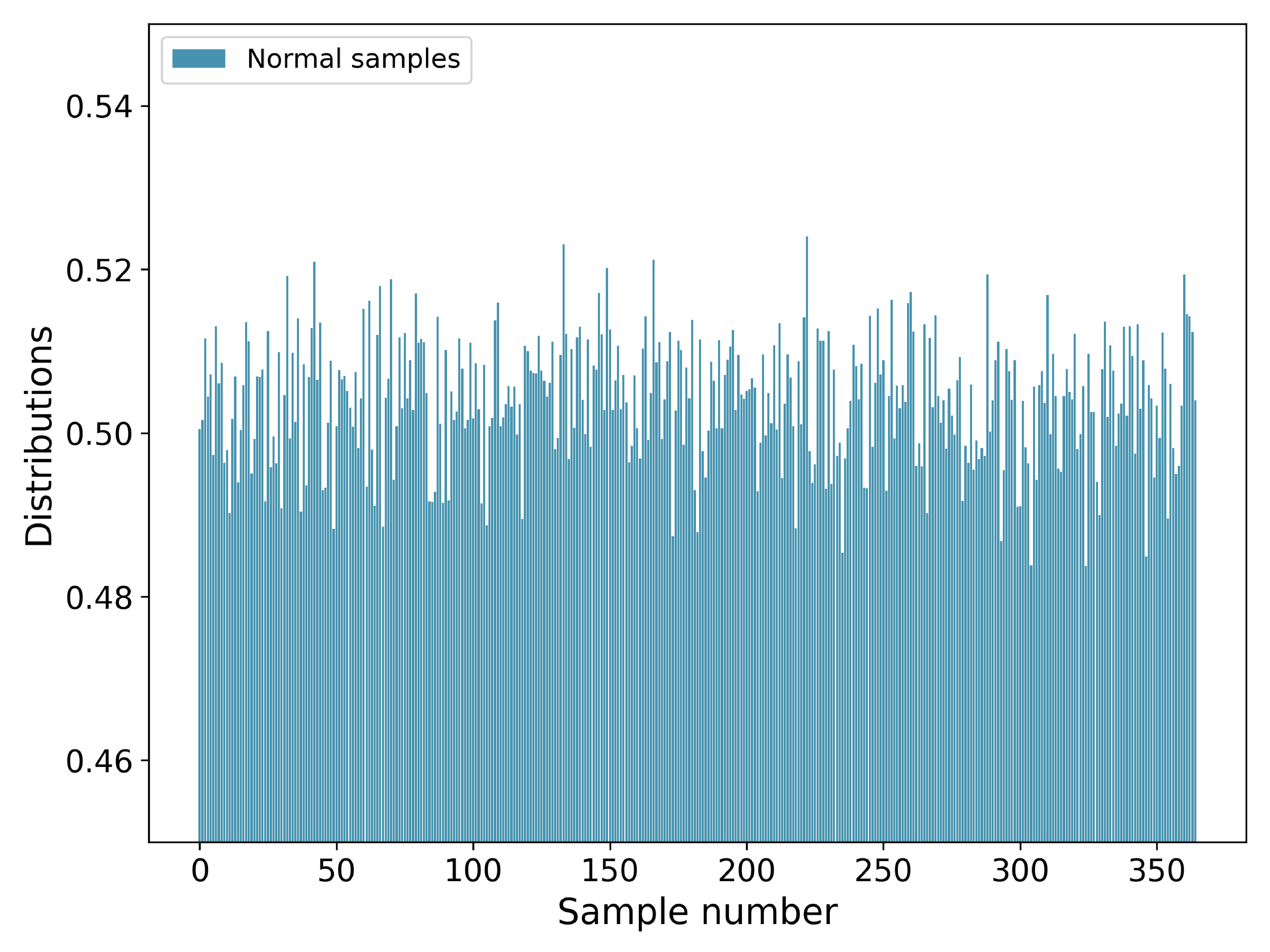}
\label{fig:Intro1}}
\subfigure[Distributions on the generated abnormal samples]{
\includegraphics[width=0.3\textwidth]{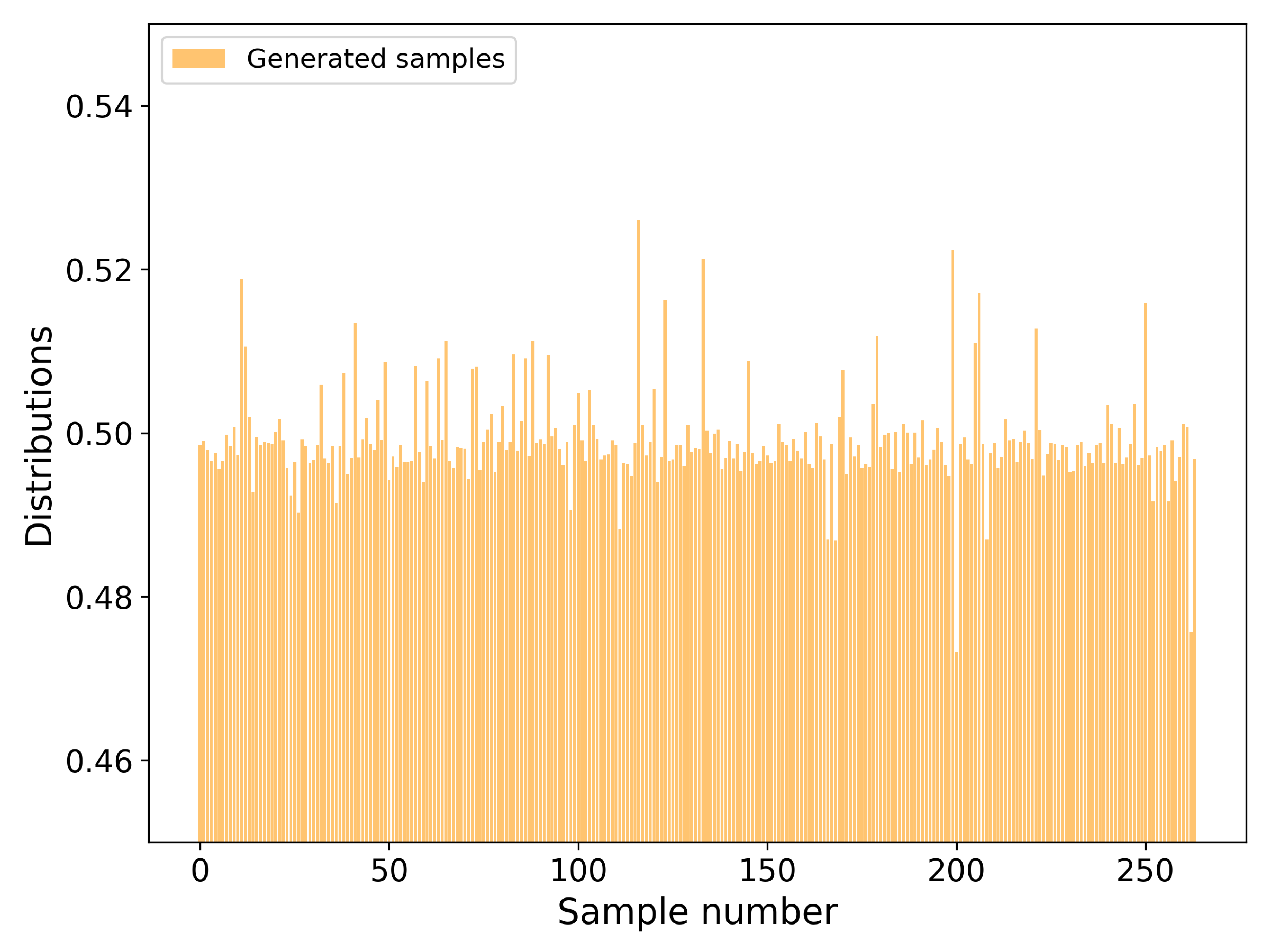} 
\label{fig:Intro2}}
\subfigure[Distributions on the abnormal samples]{
\includegraphics[width=0.3\textwidth]{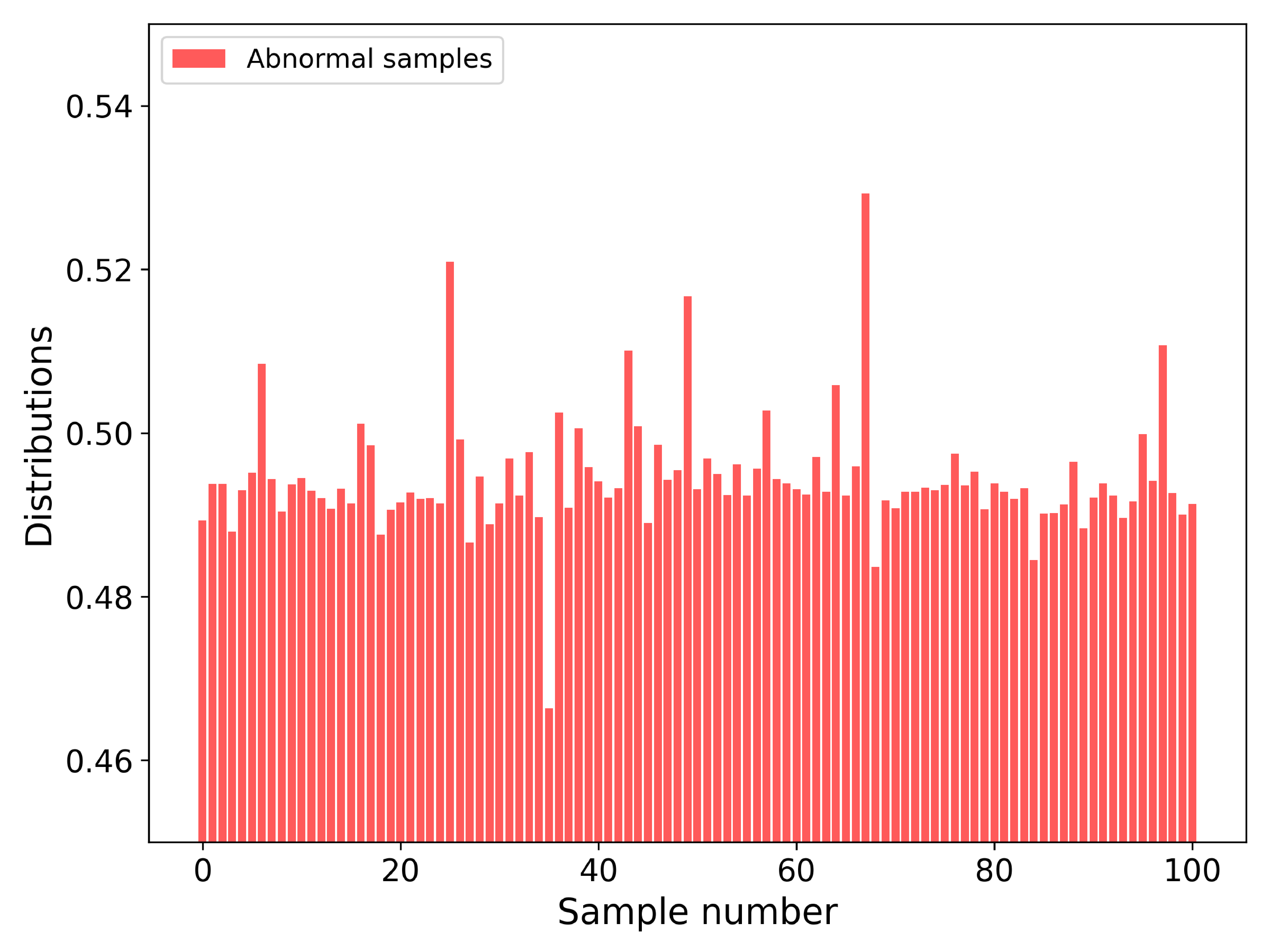} 
\label{fig:Intro3}}

	\caption{The graph-level representation distributions of the normal samples, abnormal samples, and generated samples from counterfactual theory on COX2 dataset \cite{pang2019deep,morris2020tudataset}.} \label{fig_intro}
\end{figure*}

Graph-level anomaly detection (GLAD) has been increasingly applied across a diverse array of downstream tasks to pinpoint anomalous information in numerous graphs, these tasks focus on identifying graphs that exhibit anomalous patterns within a set of graphs. For example, identifying toxic targets from protein profiles with complex structures\cite{luo2022deep,lin2023discriminative}, detecting fraudulent groups in a large number of financial transaction networks \cite{huang2022dgraph,9020760,8325544}, filtering out malicious or fake messages from a large base of user messages on social media \cite{khattar2019mvae,shu2017fake,8862186}. Besides, by processing and abstracting magnetic resonance imaging appropriately, GLAD can be applied to predicting disease \cite{luo2024graph,8862186}. For instance, Li. \textit{et al.} first construct brain graphs from neuroimaging data of subjects and then utilize GLAD methods to recognize anomalous brain graphs for disorder detection \cite{li2021braingnn}. Thus, GLAD possesses immense value and can be used for various applications. By leveraging effective GLAD methods, large volumes of data with complex structures can be effectively resolved, and a basis for relevant decision-making can be provided for the user.

The past few years have witnessed that multiple methods on GLAD have already been proposed to capture anomalous features and improve detection performance. Existing GLAD works are summarized into two following classes:  Firstly, traditional GLAD methods mainly rely on graph classification algorithms, such as isolation forest (IF) \cite{liu2008isolation}, local outlier factor (LOF) \cite{breunig2000lof}, and one-class support vector machine (OCSVM) \cite{scholkopf1999support}. For some relatively simple datasets, these methods can achieve an acceptable recognition accuracy by leveraging effective distance metrics, such as the family of graph spectral distances (FGSD) which leverages the multiset of node pairwise distances and is able to capture the unique, stable properties of the graphs \cite{verma2017hunt}. However, such traditional algorithms are often overly sensitive to the distribution of normal and abnormal samples due to their simple architecture. Furthermore, the methods mentioned are composed of two independent components: a feature extractor and a classification algorithm, collectively referred to as two-stage methods. The classification algorithm relies on the provided graph embedding features from the feature extractor which produces "hand-crafted" features \cite{zhao2023using}. Such features limit their potential for improvement.

Secondly, methods based on deep learning mainly utilize graph neural networks (GNNs) \cite{wu2020comprehensive} to learn abnormal graph representations to distinguish normal and abnormal graphs. The GNNs can effectively learn the interconnections of nodes and obtain the global structure of each graph from datasets with complex topologies and rich relational information. Multiple iterations allow for constant node representation updates, capturing complex graph relationships. Recently, the GNN-based methods have yielded more results. SIGNET \cite{liu2024towards} is an unsupervised learning method that leverages a multi-view subgraph information bottleneck framework to achieve interpretable and decent results. GLocalKD \cite{ma2022deep} adopts a knowledge distillation approach, distilling representations from graph-level and node-level. GLADST \cite{lin2023discriminative} utilizes a dual-students-teacher model to maximize the difference between normal and abnormal samples. And GLADD \cite{lin2023multi} combines multi-representations space separation to detect anomaly graphs. Although these deep learning methods can tackle GLAD tasks with acceptable efficiency and accuracy, limitations still remain. While these methods present their own unique solutions to the GLAD tasks, there are some flaws, they either fail to efficiently construct balanced and efficient datasets, or overlook fully leveraging the information from the graphs.

After analyzing the existing solutions to GLAD tasks, it is evident that the imbalance in datasets is a prevalent issue. Previous methods tend to either overlook this severe imbalance or lack effective methods to construct reliable samples. In other words, since abnormal samples are typically in the minority in GLAD tasks, training models directly on imbalanced data may cause the models to predominantly learn patterns from normal graphs, thus increasing the likelihood of misclassifying targets as normal graphs. Furthermore, if the generated abnormal samples from improper methods are not reliable enough, they might even be counterproductive. Additionally, common deep learning approaches often only consider the node features themselves, neglecting the degree attributes. These methods fail to fully utilize all valuable information from the sample graphs, leading to suboptimal model performance. To resolve the aforementioned issues, several challenges arise. Firstly, the most straightforward approach for balancing the datasets is to employ anomaly sample augmentation methods \cite{you2020graph} or direct random sampling for oversampling. However, samples generated by such methods may contain false samples which are incorrectly marked, or fail to enrich the sample features sufficiently, leading to unreliable samples. Secondly, maximizing the utilization of rich node information is the key to feature learning. Previous methods either focus solely on node features or substitute node features with degree attributes for training, resulting in suboptimal performance on feature-rich datasets. Thus exploring ways to enable node features and degree attributes to complement each other, thereby constructing more enriched and comprehensive information, is vital and meaningful.

In this work, we introduce an \textbf{I}mbalanced \textbf{G}raph-level \textbf{A}nomaly \textbf{D}etection via \textbf{C}ounterfactual augmentation and \textbf{F}eature learning (IGAD-CF) to address the mentioned issues. Firstly, to solve the problem of scarce anomalous samples, we construct an anomaly sample generation module inspired by the counterfactual theory commonly employed in contrastive learning \cite{yang2023generating}. This module can generate abnormal samples by applying appropriate perturbations to normal samples, thereby creating a balanced dataset containing both original information and manually created abnormal information. Meanwhile, it enriches the patterns of anomalous graphs, enhancing the anomaly perception ability of our model. As is shown in Figure \ref{fig_intro}, the distribution result is obtained by employing a graph convolutional network to extract feature representations from the dataset, followed by normalizing the output through a linear function to map it onto the range from 0 to 1. The figure illustrates that the generated abnormal samples bear a close resemblance to the original abnormal samples. Specifically, their distributions are quite similar, clustering stably around a certain value and generally not exceeding 0.5. In contrast, the normal samples are mostly distributed above 0.5 with greater volatility. This demonstrates that our generated abnormal samples are highly consistent with the real anomalous samples, rendering them reliable and practical. This COX2 dataset is available online\footnote{ \url{https://chrsmrrs.github.io/datasets/docs/home}}. Additionally, to facilitate comprehensive feature utilization and accurate anomaly detection within graphs, we introduce an effective node feature learning module. In this module, we introduce degree attributes to supplement the node features, thereby enriching the information representation. This enabling enhances anomaly localization and improves anomaly pattern learning capabilities through the structural information provided by degree attributes. Furthermore, we incorporate an adaptive weight learning module into the framework.  By assigning adaptive weights to different features across datasets, this module not only enhances model generalization but also enables the model to focus on features with greater impact on the results. Finally, to verify the effectiveness of the proposed framework, comprehensive experiments are designed and carried out on traditional public datasets against popular GLAD methods. Furthermore, we extend our model to practical application on brain diseases. Our model can solve problems caused by data imbalance, insufficient application of features, and individual differences in graphs. The contributions are outlined as follows:

\begin{enumerate}
    \item[$\bullet$] Firstly, we apply the anomaly sample generation module to undertake balancing procedures on unbalanced datasets to mitigate the issue of poor model generalization. By using counterfactual mechanisms, the generated abnormal samples have high reliability, which allows the model to capture more anomalous information.
    
    \item[$\bullet$] Secondly, we introduce a node feature learning module that specifically focuses on the intimate relationship between node features and degree features, utilizing the model to comprehend their inherent connection. By concatenating the convolution results, the potential interaction information between them is also taken into account while retaining their respective information.
    
    \item[$\bullet$] Thirdly, according to the fact that there are always differences between graph individuals, we train an adaptive weight learning module to pinpoint the most favorable features for detection more effectively for our model.

    \item[$\bullet$] Finally, we perform experiments on traditional public GAD datasets to confirm our model's ability and apply it to real brain-disease datasets.
\end{enumerate}

\section{Related Works}

Based on the research topic and relevant techniques involved, here, we mainly introduce graph neural networks, graph anomaly detection, and counterfactual learning to highlight the contributions of our proposed method, IGAD-CF.

\subsection{Graph Neural Networks}
For the past few years, GNNs have been successfully employed in graph anomaly detection tasks, yielding fruitful results. By leveraging the message passing mechanisms, GNNs demonstrate an enhanced capability to capture complex representations embedded within the target graphs sensitively. For node-level tasks, each node's representation is encoded with its own information and the information from the neighbors \cite{wu2020comprehensive}. Then the representation is updated according to the message vectors aggregated from neighboring nodes. A pivotal strength of this strategy is that the representation learned for each node encapsulates its own information and incorporates the information from its neighboring nodes \cite{wu2020comprehensive,gori2005new}. For example, Graph convolutional network (GCN) \cite{kipf2016semi} can use convolution operators for information aggregation to update the representation of nodes through local information. And graph attention network (GAT) \cite{velivckovic2017graph} uses attention mechanisms to assign different weights to neighboring nodes. For graph-level tasks, the node representations need to be further downsampled to obtain the overall graph representation. The common downsample methods are graph pooling operations \cite{wu2020comprehensive} which aid GNNs in extracting high-level topological information. Some improvements start with optimizing pooling operations, DiffPool, for example, is a differentiable pooling that can improve graph classification benchmarks when applied to GNNs \cite{ying2018hierarchical}. In our framework, we leverage GNNs to capture not only the node features but also the structural information of graphs, thereby enhancing the feature representations for anomalous graphs.

\subsection{Graph Anomaly Detection}
Graph anomaly detection (GAD) is commonly delineated into two genres: node-level anomaly detection (NLAD) and GLAD. 

NLAD targets identifying anomalies in the behavior of nodes within a graph by analyzing the patterns of node features or the connectivity between nodes. This method is useful in solving the problems of detecting fraudulent activities or supervising 'zombie' accounts on social networks. As an illustration, Liu. \textit{et al.} propose a method based on attention mechanisms to find out fraud transactions on e-commerce platforms \cite{liu2021intention}. Such methods based on attention mechanisms are frequent on NLAD tasks \cite{wang2019semi,tang2022rethinking}. Tang. \textit{et al.} accomplish GLAD tasks from the spectral domain and successfully achieve excellent results on datasets with a vast multitude of nodes and edges \cite{tang2022rethinking}. However, NLAD focuses on individual nodes but overlooks the global patterns from the entire graph.

GLAD strives to pinpoint anomalous graphs in a series of graphs. The graph-level anomalies may arise due to irregularities in topological structures and node behaviors, or the relationships represented by the edges. The traditional methods for GLAD employ a two-stage approach to capture anomalies at the graph level. For instance, employing the FGSD \cite{verma2017hunt} to generate spectral distance histogram vector representations for each graph, which are then fed into conventional binary classification detection techniques such as IF, LOF, and OCSVM \cite{liu2008isolation,breunig2000lof,scholkopf1999support}. These two-stage methods, with mutually independent phases, impose limitations on the model's ability to capture intricate features effectively. Consequently, these conventional methods exhibit suboptimal performance when confronted with datasets exhibiting complex features. Besides, Zhao. \textit{et al.} observe a performance flip when there is a significant imbalance in the datasets, with the detection efficacy heavily contingent on which class is sampled as the anomaly \cite{zhao2021glod-issues}. To enhance the performance of the GLAD model, recent studies leverage deep learning techniques with GNNs to capture anomalous patterns more effectively. GLocalKD \cite{ma2022deep} employs GNNs to learn graph and node representations, then performs stochastic knowledge distillation by minimizing the prediction error between these representations. OCGTL \cite{qiu2022raising} utilizes multiple GNNs to learn graph embeddings separately, achieving effective GAD by generating different perspectives of graph embeddings based on varying parameters. Although these GNN-based approaches significantly boost model performance compared to the traditional methods and, to a certain extent, improve the performance degradation caused by imbalanced datasets, they do not fundamentally address the issue of imbalanced datasets nor fully exploit the potential of GNNs to extract optimal feature representations. Thus, there remains room for improvement in these aspects.

\subsection{Counterfactual Learning}

The concept of counterfactual was initially introduced in the field of psychology \cite{kahneman1981simulation,lewis1973counterfactuals,lewis2013counterfactuals,pearl2018book}. Counterfactual describes events that do not really exist and contradict factual world knowledge \cite{kulakova2013processing,li2020survey}. In other words, counterfactual thinking involves considering a false premise about an event that has already occurred, leading to a conclusion that is contrary to the established facts \cite{stepin2021survey}. For example, if the established fact is that a worker missed the subway to work, the counterfactual would be, "If he had arrived earlier, he could have caught the subway."  By varying certain preconditions, the result may be different from the actual facts. As a result, more and more scholars in artificial intelligence have considered applying this psychological pattern to their research. This approach can improve the models' interpretability to a certain extent \cite{fernandez2019relevance,hendricks2018grounding,stepin2021survey}. The interpretability can help the researchers understand the reasons for the decision of their model \cite{byrne2019counterfactuals}. Furthermore, Chen. \textit{et al.} utilize counterfactual theory to process image datasets in the visual question answering domain, augmenting minority samples and capturing the rich inherent information of the majority classes \cite{chen2023counterfactual}. Since counterfactual theory involves altering conditions to influence outcomes \cite{mcgill1993contrastive}, Yang. \textit{et al.} employ counterfactual theory to perturb the features of graphs, generating samples with features similar to the original samples but with opposite semantic labels, enabling contrastive learning \cite{yang2023generating}. This method not only preserves the original samples' information but also identifies the most discriminative features that largely determine the sample labels through feature perturbation. In our work, we employ counterfactual learning to generate anomalous samples to address the issue of imbalanced anomaly samples. Since counterfactual learning aims to generate samples that are similar to normal samples but with opposite labels, it can supplement the scarce anomaly samples.

\section{Problem Definition}
The given dataset $\mathcal{G}$ = \{$G_1$, $G_2$ ..., $G_N$\} consists of two mutually disjoint subsets: the set of normal samples $\mathcal{G}_{nor}$ and the set of abnormal samples $\mathcal{G}_{abn}$. Thus the relation of these sets is: $\mathcal{G} = \mathcal{G}_{nor} \cup \mathcal{G}_{abn}$. Furthermore, $\mathcal{G}_{abn}$ can be divided into two subsets based on whether the abnormal samples are original anomalies $\mathcal{G}_{ori}$ or generated anomalies $\mathcal{G}_{gen}$. Similarly, we have $\mathcal{G}_{abn} = \mathcal{G}_{ori} \cup \mathcal{G}_{gen}$. For every graph $G_i$ = \{${V}_{i}$, ${E}_{i}$, ${D}_{i}$, ${X}_{i}$, ${A}_{i}$\} $\in \mathcal{G}$, ${V}_{i}$ = \{$v_1$, $v_2$, ..., $v_n$\} represents the nodes in the corresponding graph, and ${E}_{i}$ =\{$e_1$, $e_2$, ..., $e_m$\} is the set of edges connecting the nodes. ${D}_{i}$ =\{$d_1$, $d_2$, ..., $d_n$\} $\in \mathbb{R}^{n \times 1}$ donates degree attributes that describe the number of edges connected to each node, ${X}_{i}$ = \{$x_1$, $x_2$, ..., $x_n$\} $\in \mathbb{R}^{n \times h}$ corresponds to the matrix of node features. ${A}_{i}$ $\in \mathbb{R}^{n \times n}$ denotes the adjacency matrix, which is typically binary, with the value of 1 indicating the presence of an edge. The samples generated by counterfactual learning are similar to normal samples but have opposite labels, i.e., generated abnormal samples $\mathcal{G}_{gen}$. Based on the dataset $\mathcal{G}$, we aim to complete GLAD tasks by constructing a model that can calculate each graph's final output score $O$. If $O$ is closer to 1, the input graph is more likely to be abnormal, while it is closer to 0, the input could be normal.

\section{Method}

\begin{figure*}[h!]
	\small
	\includegraphics[width=\textwidth]{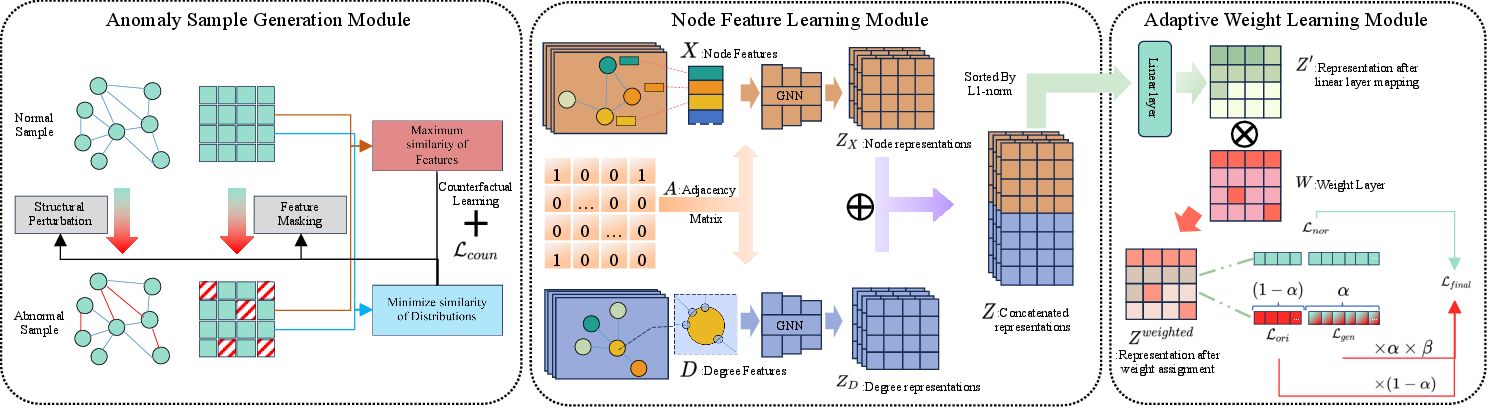}
	\caption{The framework of the IGAD-CF. The anomaly sample generation module employs counterfactual learning to perturb the adjacency matrices and feature matrices of normal samples. This module guarantees the efficacy and reliability of the generated samples. The node feature learning module combines the node features and degree attributes, capturing richer and more complete information. The adaptive weight learning module utilizes an adaptive weight matrix, ensuring that each feature is assigned a weight commensurate with its importance.} \label{fig1}
\end{figure*}

In this part, we will deliver a detailed delineation of the methodology for the IGAD-CF framework. The overall framework is shown in Figure \ref{fig1}. First, we will detail the process of utilizing an anomaly sample generation module to supplement the imbalanced dataset. Next, an introduction to how we extract more effective node features and assign them adaptive weights will be shown. Finally, we will present the delineation of anomalous graphs and introduce a complete loss function with hyperparameters.

\subsection{Anomaly Sample Generation Module}

In the task of GLAD, anomalous samples from the dataset typically represent the minority class. This implies that the majority of effective information in the dataset represents normal samples, leading to a lack of anomalous information. The absence of anomalous features in large datasets makes it challenging for machine learning methods to learn rich features, resulting in poor generalization and recognition capabilities. Therefore, generating anomalous samples with abundant anomalous information to supplement the original data is crucial. The CGC \cite{yang2023generating} method leverages counterfactual theory to perturb given samples, generating samples similar to the originals but with opposite labels. Inspired by the CGC method, we employ this approach to generate anomalous samples. Since our problem is essentially a binary classification task, generating samples with opposite labels from the majority class can effectively supplement the minority class. Moreover, this method can construct samples that retain rich information, enabling the generation of efficient and near-realistic abnormal samples. The core of this generation method is to obtain two perturbation matrices for modifying the structures and node features of normal graphs. Specifically, we apply structural perturbation matrix $M_a$ and feature masking matrix $M_b$ to every single graph in the selected set $\mathcal{G}_{sel} \subseteq \mathcal{G}_{nor}$, the number of $\mathcal{G}_{sel}$ is equal to the gap between $\mathcal{G}_{nor}$ and $\mathcal{G}_{ori}$. $G_j$ represents the $j$-th graph in $\mathcal{G}_{sel}$. We initialize a trainable adjacency matrix $M_a$, which is a perturbation matrix trained under the guidance of every adjacency matrix $A_j$ of $G_j$. During training, we multiply $M_a$ with the $A_j$ and then apply a threshold function to binary the result into a new 0/1 adjacency matrix $A_j'$.  
\begin{equation}
A_j' = {I}(sigmoid(M_a \times A_j) \geq \sigma).\label{CGCeq1}
\end{equation}
The function $I(\cdot) $ here donates an indicating function that returns 0 or 1 depending on the threshold $\sigma$.
Next, we initialize another trainable feature masking matrix $M_b$, which is also produced via a threshold function depending on the threshold $\tau$ to obtain a 0/1 mask matrix $M_b'$.  We applied an element-wise multiplication of the binary mask $M_b'$ and the feature matrix $X_j$ from $G_j$ to obtain the masked feature matrix $X_j'$.  In other words, $X_j'$ is obtained by taking the Hadamard product of  $M_b$ and $ X_j$. The process can be formalized as:
\begin{equation}
M_b' = {I}(sigmoid(M_b) \geq \tau),\label{CGCeq2}
\end{equation}
\begin{equation}
X_j' = M_b' \odot X_j.\label{CGCeq3}
\end{equation}
Then, the overall loss comprises two principal subparts based on the counterfactual reasoning mechanism: maximizing similarity to the original samples and enforcing dissimilarity in labels. We first calculate the Frobenius norm between the original adjacency matrix $A_j$ and the perturbed $A_j'$ and that of the mask matrix $M_b'$. Then we combine them as the first part of the overall loss during the generation process $\mathcal{L}_{G1}$:
\begin{equation}
\mathcal{L}_{G1} = ||A_j - A_j'||_F -||M_b'||_F, \label{CGCeq4}
\end{equation}
where the function calculates the Frobenius norm inside. Subsequently, we evaluated the probability distributions of the original and perturbed graphs using the KL-Divergence function $D_{KL} (\cdot) $.  This builds the second part of the overall loss $\mathcal{L}_{G2}$.
\begin{equation}
\mathcal{L}_{G2} = D_{KL}(p,p_a) + D_{KL}(p,p_b). \label{CGCeq5}
\end{equation}
Finally, the two loss terms were combined to satisfy the counterfactual reasoning mechanism, thereby we get the final loss of this generation module $\mathcal{L}_{coun}$.
\begin{equation}
\mathcal{L}_{coun} = \mathcal{L}_{G1} - \mathcal{L}_{G2}. \label{CGCeq5}
\end{equation}
After training, we applied the learned perturbation matrix $M_a$ and mask matrix $M_b$ simultaneously to the sample $G_j$, perturbing both node features and edges, to supply the minority to the same size as the majority.

Thus, the two classes can have the same number of training samples, which can help avoid model bias when classifying positive and negative samples in training. So that our model can better capture the features from normal and anomalous samples, and improve the overall performance on detecting anomaly, so as not to bias the class that is dominant in data volume. In short, this module contributes to enhancing the overall generalization and performance of our model. 

\subsection{Node Feature Learning Module}

Existing GCN methods may focus more on the impact of node features themselves on node anomalies \cite{lin2023discriminative} while overlooking certain structural factors. In our experiments, we find that when using node features for anomalous graph learning, incorporating degree attributes as corresponding structural information into the model can complement the information learned by the model, enabling it to learn more comprehensive and effective anomalous information. This approach can effectively pinpoint the feature information of anomalies. When constructing the Node Feature Learning Module, we use the method of combining the node feature and the degree feature, making the features complementary.

To fully leverage node features and degree features, as well as uncover potential relationships between them, we capture these features separately after data balance. Thereafter, we concatenate them in a principled manner to capture their intrinsic connections, thereby enhancing the accuracy of the GAD task. To be exact, during the data preprocessing stage, we constructed two sets of input features: Node attribute feature matrices $X$ and degree attributes $D$. To utilize $X$ and $D$ better, we designed an encoder model based on GCNs. This model first encodes the node features $X$ and degree attributes $D$ separately to get corresponding representations from node features $Z_X$ and that from degree attributes $Z_D$:
\begin{equation}
    Z_X=GCN_{X}(X, A),\label{eq8}
\end{equation}
\begin{equation}
    Z_D=GCN_{D}(D, A),\label{eq9}
\end{equation}
where $A$ is the graph adjacency matrices, encoding the aggregated attribute and degree features of neighboring nodes, respectively. Next, $Z_X$ and $Z_D$ are concatenated in the last dimension as:
\begin{equation}
    Z=Concat(Z_X , Z_D,dim = -1),\label{eq10}
\end{equation}
where $Concat(\cdot)$ represents the operation of concatenating two matrices on the $dim^{th}$ dimension, and $Z$ denotes the concatenated feature that fuses the node attribute and structural topology information. Thus $Z$ is capable of capturing the intrinsic associations between them.

\subsection{Adaptive Weight Learning Module}

Given that, each graph has multiple feature information, and different feature information holds varying degrees of importance for the model's training. We introduce an Adaptive Weight Learning Module $W$, which can adaptively focus on the information with higher assigned weights. $W$ can aid in determining which features can help the model make predictions to a greater extent. Such appropriate weight allocation becomes particularly important after combining the multiple node features and degree features. Specifically, we first sort the obtained feature $Z$ based on the L1 distance $d_i$ of each node to the origin:
\begin{equation}
    di = ||x_i||_{1},\forall i \in \{1,2,...,n\},\label{eq11}
\end{equation}
where the function calculates the L1 norm of each node $x_i$. Then, we sort $Z$ in descending order based on $di$, obtaining the sorted feature $Z^{sorted}$:
\begin{equation}
    Z^{sorted}=Z[:,sort(-d),:], d=[d_1,d_2,...,d_n].\label{eq12}
\end{equation}
This is to prioritize the processing of node vectors that are farther away from the origin, as these node features are more prominent, making it more likely for them to be assigned higher weights in the subsequent weight allocation. Then we map $Z^{sorted}$ with a linear layer to reduce its dimension and get $Z'$. Finally, we multiply $Z'$ by the weight matrix $W$, obtaining a feature $Z^{weighted}$ that is adaptively weighted.
\begin{equation}
    Z^{weighted}=Z' \times W.\label{eq_getZW}
\end{equation}
\subsection{Loss Function }
After obtaining the weighted feature, we utilize an effective loss function for our model. Before feeding the balanced samples to the model, we divide the normal and abnormal samples, with the first half being normal samples and the others being abnormal samples. Based on the division, we split the loss after each forward passes into the loss $\mathcal{L}_{nor}$ from normal samples and the loss $\mathcal{L}_{abn}$ from all abnormal samples, including the original abnormal samples and the generated abnormal samples. $\mathcal{L}_{nor, i}$ and $\mathcal{L}_{abn, i}$ represent the $i$-th loss of normal or abnormal samples, $N_{nor}$ and $N_{abn}$ donate the number of normal samples and the number of all abnormal samples. To this end, we first construct an initial loss $\mathcal{L}_{initial}$ formula: 

\begin{equation}
\mathcal{L}_{nor} = -\frac{1}{N_{nor}} \sum_{i=1}^{N_{nor}} \log(1 - \mathcal{L}_{nor, i}),
\end{equation}
\begin{equation}
\mathcal{L}_{abn} = -\frac{1}{N_{abn}} \sum_{i=1}^{N_{abn}} \log(\mathcal{L}_{abn, i}),
\end{equation}
\begin{equation}
    \mathcal{L}_{initial} = \mathcal{L}_{nor} +\mathcal{L}_{abn}\label{eq1}.
\end{equation}
Benefiting from the anomaly sample generation module, these two-class samples are well balanced. However, considering that the generated abnormal samples used are not from the original data, we introduce a hyperparameter $\beta$ to limit the influence of these generated abnormal samples during training. 
Furthermore, the weights assigned to generated abnormal samples in the final loss calculation should be determined by their proportion within all abnormal samples. Thus, we first calculate the ratio $\alpha$ of generated abnormal samples $N_{gen}$ to the number of all anomaly samples $N_{abn}$. After that, we divide the loss of all anomaly samples into two parts according to $\alpha$.
\begin{equation}
    \alpha = N_{gen} / N_{abn}.\label{eq2}
\end{equation}
We multiply the loss from generated abnormal samples by the hyperparameter $\beta$. This way, we achieve manual control over the degree of influence of generated abnormal samples on the model. The $\mathcal{L}_{ori}$ represents the loss from the original abnormal samples and $\mathcal{L}_{gen}$ represents the loss from the generated abnormal samples, $N_{ori}$ and $N_{gen}$ donate the number of original samples and the number of generated abnormal samples, and $\mathcal{L}_{ori, i}$ and $\mathcal{L}_{gen, i}$ represent the $i$-th loss of original or generated abnormal samples.
\begin{equation}
\mathcal{L}_{ori} = -\frac{1}{N_{ori}} \sum_{i=1}^{N_{ori}} \log(\mathcal{L}_{ori, i}),
\end{equation}
\begin{equation}
\mathcal{L}_{gen} = -\frac{1}{N_{gen}} \sum_{i=1}^{N_{gen}} \log(\mathcal{L}_{gen, i}).
\end{equation}
Thus, $\mathcal{L}_{abn}$ is decomposed into two components, $\mathcal{L}_{gen}$ and $\mathcal{L}_{ori}$, such that $\mathcal{L}_{abn}$ = $\mathcal{L}_{ori}$ $\times$ (1 - $\alpha$) + $\mathcal{L}_{gen}$ $\times$ $\beta$ $\times$ $\alpha$. The initial formula is upgraded to calculate the final loss $\mathcal{L}_{final}$:
\begin{equation}
    \mathcal{L}_{final} = \mathcal{L}_{nor} + \mathcal{L}_{ori}\times(1-\alpha) + \mathcal{L}_{gen}\times\beta\times\alpha.
\end{equation}
\begin{table*}[!h]
\centering
\caption{The summary of experimental datasets.}
\centering
\begin{tabular}{c|ccccc}
\hline
Datasets    & Graphs & Avg-nodes & Avg-edges & Normal Instances & Abormal Instances \\ \hline
AIDS        & 2,000   & 15.69     & 16.2      & 391              & 1,609              \\
BZR         & 405    & 35.75     & 38.36     & 327              & 78                \\
COX2        & 467    & 41.22     & 43.45     & 365              & 102               \\
DHFR        & 756    & 42.43     & 44.54     & 473              & 283               \\
ENZYMES     & 600    & 32.63     & 62.14     & 160              & 440               \\
IMDB-BINARY & 1,000   & 19.77     & 96.53     & 600              & 400               \\
NCI1        & 4,110   & 29.87     & 32.3      & 2,464             & 1,646              \\
hERG        & 655    & 26.48     & 28.79     & 164              & 491               \\ \hline
\end{tabular}
\label{tab:datasets}
\end{table*}
\subsection{Graph Anomaly Detection Module}
After completing the model's training, the test datasets are applied to evaluate the model's accuracy. In our graph anomaly detection module, we first apply the node feature learning module to the received test data to handle the adjacency matrix, node features, and degree attributes. Subsequently, the Adaptive Weight Learning Module is employed to the results from the previous step, obtaining the $Z^{weighted}$. Then, a linear transformation maps the results to a single value, which is normalized to the range of 0 to 1. Specifically, we first applied a non-linear transformation to the obtained $Z_R$ using the $ReLU (\cdot)$ function. 
\begin{equation}
    Z_R=ReLU(Z^{weighted}).\label{eq_res1}
\end{equation}
Subsequently, we use a fully connected layer for processing a result $U$, mapping the features to a scalar value.
\begin{equation}
    U=W_dZ_R + b_d.\label{eq_res2}
\end{equation}
In this process, $W_d$  represents the weight and $b_d$ is the bias of the fully connected layer. Finally, we utilized the Sigmoid function $\sigma(\cdot) $ to constrain the output $O$, ensuring the result falls within the range of 0 to 1, facilitating the interpretation of whether the input graph is normal or not.
\begin{equation}
    O=\sigma(U).\label{eq_res3}
\end{equation}
Then we apply a heaviside step function $H(\cdot)$ to the output $O$, when the result exceeds the certain threshold $t$, it indicates a higher likelihood of the graph being anomalous, and vice versa. In our approach, we set the default threshold to 0.5.
\begin{equation}
H(O-t) = \begin{cases}
1, & \text{if } O-t > 0;\\
0, & \text{otherwise.}
\end{cases}\label{eq_res4}
\end{equation}

\section{Experiments}
\subsection{Dataset}
In order to assess the efficacy of our framework on datasets with higher recognition, we selected 8 real-world anomaly detection datasets from various fields \cite{pang2019deep,morris2020tudataset}. Initially, we chose commonly used datasets for graph anomaly detection, such as AIDS, BZR, COX2, DHFR, and NCI1. These are typical datasets of small molecules with relatively smaller amounts of graphs, usually constructed based on abstract atoms of each molecule as nodes, and abstract chemical bonds between atoms as edges.
Additionally, we selected the ENZYMES dataset from bioinformatics, where the graphs represent proteins, with amino acids as nodes and edges constructed based on their spatial distances and states. Furthermore, to verify the situation in actual social networks, we used the IMDB-BINARY dataset for experiments, where each independent individual is abstracted as a node, and edges are added to connect any two individuals with a collaborative relationship. Finally, we included the hERG dataset from biological joint molecules to test our model’s performance. The number of graphs, the average number of edges, and the average number of nodes in these datasets are shown in Table \ref{tab:datasets}. When preprocessing the datasets, we select these data samples which are labeled 1 as the anomalous instances, and the rest as normal instances. During the data balancing process, we perform sample generation on the minority class samples within the dataset. For some datasets (e.g., AIDS, ENZYMES, hERG) where abnormal samples outnumber normal samples, we similarly employ the counterfactual approach to construct normal samples, thereby balancing the datasets.

\subsection{Baselines}
To showcase that our proposed method is not only an improvement over traditional detection methods but also shows advantages over newer methods, we first choose the traditional methods based on conventional binary classification detection techniques:

\begin{enumerate}
    \item[$\bullet$]\textbf{FGSD-LOF} is a traditional anomaly detection approach that utilizes a two-stage method, which is composed of two parts, the FGSD \cite{verma2017hunt} algorithm to calculate the distances between node vectors and the LOF \cite{breunig2000lof} to classify the dataset basing on the distances. LOF detects the abnormal by comparing each instance's local density to the neighbors.
    \item[$\bullet$]\textbf{FGSD-IF} has a similar structure, while the detector is replaced with IF \cite{liu2008isolation}, which makes isolating observation by random feature selection and value splitting.
    \item[$\bullet$]\textbf{FGSD-OCSVM} is driven by the FGSD distance algorithm as well. Its detection core OCSVM \cite{scholkopf1999support} identifies the anomalies by their distance from the hyperplane.
\end{enumerate}
Then we select some deep learning methods, to exhibit the superiority of IGAD-CF: 
\begin{enumerate}
    \item[$\bullet$]\textbf{OCGTL} \cite{liu2024towards} is a method that has demonstrated excellence in graph anomaly detection within the domains of bioinformatics, small molecules, and social networks. OCGTL absorbs the strengths of one-class classification and transfer learning, achieving graph embeddings from multiple perspectives by integrating various GNNs, thereby effectively utilizing deep learning for GAD tasks.
    \item[$\bullet$]\textbf{GLocalKD} \cite{ma2022deep} is a GCN-based method which employs the concept of stochastic knowledge distillation, adeptly capturing both local and global information in graphs. 
    \item[$\bullet$]\textbf{GOOD-D} \cite{liu2023good} is a representative of unsupervised contrastive learning, which captures the latent patterns of in-distribution (ID) graphs by contrastive learning.
    \item[$\bullet$]\textbf{SIGNET} \cite{liu2024towards} is an explainable unsupervised learning method based on a Multi-View Subgraph Information Bottleneck. It can capture the substructures to predict the agreement between two views.
    \item[$\bullet$]\textbf{GLADST} \cite{lin2023discriminative} is constructed of dual-students-teacher model which can differentiate between normal and abnormal samples to a greater extent.
\end{enumerate}

\begin{table*}[!h]
\caption{The anomaly detection performance in terms of mean value of AUC (\%) and standard deviation.}
\resizebox{\linewidth}{!}{

\begin{tabular}{c|ccccccccc}
\hline Datasets & FGSD-LOF    & FGSD-IF     & FGSD-OCSVM  & OCGTL \cite{liu2024towards}             & GlocalKD \cite{ma2022deep}   & GOOD-D \cite{liu2023good}       & SIGNET \cite{liu2024towards}      & GLADST \cite{lin2023discriminative}              & IGAD-CF             \\ \hline
AIDS        & 12.84\tiny±5.02 & 0.60\tiny±0.89  & 16.47\tiny±6.07 & 83.49\tiny±3.78        & 99.32\tiny±0.66 & 44.46\tiny±15.73 & 99.51\tiny±0.39  & \textbf{99.71\tiny±0.59} & {\underline{99.71\tiny±0.60}}    \\
BZR         & 50.44\tiny±8.73 & 55.49\tiny±6.06 & 58.85\tiny±6.44 & {\underline{83.81\tiny±5.82}}  & 63.44\tiny±9.04 & 65.68\tiny±11.10 & 51.54\tiny±11.52 & 83.05\tiny±3.94          & \textbf{87.95\tiny±2.62} \\
COX2        & 43.29\tiny±4.85 & 43.51\tiny±3.45 & 45.77\tiny±5.81 & 61.62\tiny±5.11        & 62.32\tiny±5.14 & 58.60\tiny±5.78  & 65.47\tiny±4.80  & {\underline{72.52\tiny±7.65}}    & \textbf{77.68\tiny±3.73} \\
DHFR        & 50.80\tiny±5.94 & 48.38\tiny±5.25 & 44.11\tiny±4.45 & 61.66\tiny±5.35        & 72.07\tiny±4.49 & 76.70\tiny±2.40  & 75.46\tiny±2.80  & {\underline{81.72\tiny±1.84}}    & \textbf{87.78\tiny±1.95} \\
ENZYMES     & 61.31\tiny±3.04 & 51.24\tiny±4.55 & 57.97\tiny±7.52 & {\underline{72.43\tiny±10.32}} & 39.16\tiny±4.41 & 54.00\tiny±7.61  & 51.06\tiny±2.63  & 61.67\tiny±4.97          & \textbf{80.16\tiny±3.09} \\
IMDB-BINARY & 45.69\tiny±5.01 & 49.43\tiny±3.9  & 54.95\tiny±3.07 & 59.34\tiny±5.53        & 41.22\tiny±4.71 & 60.80\tiny±3.68  & 56.72\tiny±5.29  & {\underline{63.14\tiny±2.63}}    & \textbf{77.99\tiny±1.07} \\
NCI1        & 40.50\tiny±1.26 & 64.15\tiny±0.91 & 52.24\tiny±2.23 & 63.90\tiny±1.47        & 42.80\tiny±1.24 & 68.01\tiny±1.73  & 57.34\tiny±7.63  & {\underline{68.42\tiny±1.68}}    & \textbf{72.15\tiny±1.46} \\
hERG        & 59.61\tiny±4.61 & 63.32\tiny±5.50 & 41.27\tiny±3.50 & 60.67\tiny±11.58       & 71.89\tiny±5.73 & 71.09\tiny±3.78  & 68.30\tiny±1.21  & {\underline{73.94\tiny±6.18}}    & \textbf{75.67\tiny±4.31} \\ \hline
\end{tabular}
}
    \label{tab:Performance}
\end{table*}
\begin{table*}[!h]
\centering
\caption{The mean value of AUC (\%) and standard deviation of IGAD-CF and variant models.}
\begin{tabular}{c|ccccccc}
\hline Datasets & IGAD-CF              & w/o ASGM   & w/o AWLM   & w/o $GCN_{D}$          & w/o $GCN_{X}$     & w/o $\mathcal{L}_{nor}$       & w/o $\mathcal{L}_{abn}$ \\ \hline
AIDS & 99.71\tiny±0.60          & 99.70\tiny±0.60  & 99.70\tiny±0.60  & \textbf{99.72\tiny±0.56} & 99.70\tiny±0.61    & 90.62\tiny±4.73    & 99.67\tiny±0.66       \\
BZR  & \textbf{87.95\tiny±2.62} & 85.66\tiny±2.31 & 87.95\tiny±1.44 & 86.42\tiny±2.07          & 66.79\tiny±2.44          & 50.00\tiny±0.00  & 50.00\tiny±0.00       \\
COX2 & \textbf{77.68\tiny±3.73} & 76.56\tiny±4.53 & 76.81\tiny±5.32 & 77.18\tiny±4.30          & 66.05\tiny±5.80         & 50.00\tiny±0.00   & 54.29\tiny±5.38      \\
DHFR & \textbf{87.78\tiny±1.95} & 87.12\tiny±1.93 & 87.35\tiny±1.60 & 86.50\tiny±2.10          & 66.11\tiny±3.00          & 50.00\tiny±0.00    & 50.00\tiny±0.00     \\
NCI1 & \textbf{72.15\tiny±1.46} & 71.87\tiny±2.13 & 69.06\tiny±1.67 & 70.16\tiny±2.15          & 68.31\tiny±1.49          & 50.00\tiny±0.00  & 61.00\tiny±8.42       \\ \hline
\end{tabular}
\label{ablation}
\end{table*}

\subsection{Parameter Settings}
In our IGAD-CF, we apply two GCN layers with $d-256-128$ dimensions, where $d$ represents the dimension size of the features. The hyperparameter $\beta$ is 1.2 on all the datasets, except for BZR ($\beta$ = 0.6) and DHFR ($\beta$ = 1.4). The learning rate for most datasets is 0.001, while it is 0.0001 for AIDS and NCI1. We perform the experiments on an 11th Gen Intel Core i7-11700 CPU @ 2.50GHz, 64 GB RAM, and an NVIDIA GeForce RTX 3090 GPU. We first train the anomaly sample generation module to generate a balanced and enriched dataset. Next, we perform the node feature learning on this dataset and supplement the features with degree attributes. Following this, we apply an adaptive weight matrix to assign proper weights to the feature vectors. Finally, we optimize the model by calculating the sample loss with our proposed loss function. For the parameter settings in the baselines, we choose to use the recommended settings, which makes the comparison more convincing.

\subsection{Anomaly Detection Performance}
In order to show the robust performance of IGAD-CF on GLAD tasks across most scenarios, we conduct comparative analyses with all the mentioned baselines on the mentioned datasets. During the experimental phase, we employ a five-fold cross-validation approach to ensure the credibility of our results. Subsequently, we record the average value of AUC and standard deviation results post-testing to reflect the IGAD-CF's accuracy and stability.

The top-performing approach for each dataset is highlighted in bold font within Table \ref{tab:Performance}, while the second-best method is underlined. On the AIDS dataset, the standard deviation of IGAD-CF is slightly lower than GLADST, but the mean AUC of IGAD-CF and GLADST are both the optimal cases on this dataset. Apart from the AIDS dataset, IGAD-CF demonstrates superior performance on the remaining seven datasets, and in most of these datasets, IGAD-CF achieves an AUC improvement of at least 4\% compared to the second-best method. Notably, on the IMDB-BINARY dataset, the improvement is as high as 14.85\%, indicating that our method can capture anomalous information more effectively and classify the datasets better. Overall, it can be observed that IGAD-CF not only achieves a higher mean AUC value but also exhibits relatively low standard deviations across all datasets, with all standard deviations being less than 5\%. The largest standard deviation occurs on the hERG dataset, at only 4.31\%. This suggests that our model can more consistently achieve superior performance and stronger generalization capabilities when faced with different data groupings.

\begin{table*}[!h]
\caption{The mean value of AUC (\%) and standard deviation of the deep learning baselines when applying initial node features or employing identity methods to construct features.}
\centering
\begin{tabular}{c|cc|cc|cc}
\hline
\multirow{2}{*}{Datasets} & \multicolumn{2}{c|}{GLocalKD \cite{ma2022deep}}             & \multicolumn{2}{c|}{GOOD-D \cite{liu2023good}}                & \multicolumn{2}{c}{GLADST \cite{lin2023discriminative}}                \\ \cline{2-7} 
                          & Initial features    & Identity features   & Initial features    & Identity features    & Initial features    & Identity features   \\ \hline
AIDS                      & 95.10\tiny±1.87          & \textbf{99.32\tiny±0.66} & 14.28\tiny±7.77          & \textbf{44.46\tiny±15.73} & 97.67\tiny±0.81          & \textbf{99.71\tiny±0.59} \\
BZR                       & 62.57\tiny±7.32          & \textbf{63.44\tiny±9.04} & 29.92\tiny±9.58          & \textbf{65.68\tiny±11.10} & 81.02\tiny±3.00          & \textbf{83.05\tiny±3.94} \\
COX2                      & 62.21\tiny±5.35          & \textbf{62.32\tiny±5.14} & 42.13\tiny±1.45          & \textbf{58.60\tiny±5.78}  & 63.05\tiny±9.59          & \textbf{72.52\tiny±7.65} \\
DHFR                      & 55.05\tiny±3.58          & \textbf{72.07\tiny±4.49} & 61.61\tiny±4.84          & \textbf{76.70\tiny±2.40}  & 77.36\tiny±3.49          & \textbf{81.72\tiny±1.84} \\
NCI1                      & 31.77\tiny±1.53          & \textbf{42.80\tiny±1.24} & 34.26\tiny±2.36          & \textbf{68.01\tiny±1.73}  & 68.12\tiny±1.60          & \textbf{68.42\tiny±1.68} \\
ENZYMES                   & \textbf{47.91\tiny±6.17} & 39.16\tiny±4.41          & \textbf{54.22\tiny±4.96} & 54.00\tiny±7.61           & \textbf{69.43\tiny±9.14} & 61.67\tiny±4.97          \\ \hline
\end{tabular}
\label{feature table2}
\end{table*}
\begin{table}[!h]
\centering
\caption{The mean value of AUC (\%) and standard deviation. This table presents the performance of IGAD-CF across multiple datasets when using different types of features for training. The backslash symbol (\textbackslash{}) denotes the inability to obtain a valid or meaningful AUC result.}

\begin{tabular}{c|ccc}
\hline Datasets & Identity            & LDP                 & DB          \\ \hline
AIDS        & 99.71\tiny±0.60          & \textbf{99.77\tiny±0.46} & 99.70\tiny±0.60          \\
BZR         & 86.28\tiny±1.73          & 77.44\tiny±8.26          & \textbf{87.66\tiny±2.68} \\
COX2        & \textbf{77.68\tiny±3.73} & 61.35\tiny±5.43          & 68.31\tiny±6.45          \\
DHFR        & \textbf{87.11\tiny±2.18} & 71.14\tiny±1.20          & 77.85\tiny±2.92          \\
ENZYMES     & \textbf{80.16\tiny±3.09} & 62.76\tiny±2.92          & \textbackslash{}    \\
IMDB-BINARY & \textbf{77.99\tiny±1.07} & 56.64\tiny±1.98          & 64.56\tiny±3.10          \\
NCI1        & \textbf{72.15\tiny±1.46} & 68.99\tiny±1.70          & 71.13\tiny±1.97          \\
hERG        & 75.67\tiny±4.31          & 80.95\tiny±3.51          & \textbf{82.78\tiny±4.03} \\ \hline
\end{tabular}

\label{feature table1}
\end{table}
\subsection{Ablation Study}

Apart from carrying out baseline comparison experiments on the public datasets, we systematically deconstruct the model and conduct a series of ablation studies on diverse datasets, including AIDS, BZR, COX2, DHFR, and NCI1, to elucidate the impact of each component. First, to showcase the effectiveness of our generated abnormal samples, we delete the Anomaly Sample Generation Module (w/o ASGM) and directly train the entire model with the imbalanced datasets. Second, we omit the Adaptive Weight Learning Module (w/o AWLM) to show the improvement from proper weight assignment. Then, to demonstrate the performance optimization brought about by the complementarity of node features and degree attributes, we remove the GCN that processes node features (w/o $GCN_{X}$) and the GCN that processes degree attributes (w/o $GCN_{D}$) respectively. Finally, our losses are obtained by combining losses from abnormal samples $\mathcal{L}_{abn}$ with losses from normal samples $\mathcal{L}_{nor}$, so we eliminate each part of them one by one (w/o $\mathcal{L}_{abn}$ and w/o $\mathcal{L}_{nor}$) to ascertain the significance of each loss. For each ablation model, we remove the mentioned parts only and leave the remaining parts unchanged for targeted analysis. The result is exhibited in Table \ref{ablation}.

The results indicate that the entire IGAD-CF consistently exceeds the ablated models on these selected datasets except for AIDS in which every method produces a similar score. The suboptimal performance of our entire model on the AIDS could be attributed to the relatively small number of edges and nodes, resulting in simpler structures. In such cases, the node features or degree attributes alone may be sufficient to capture the necessary information, leading to generally good performance across ablation models. Moreover, the highest average AUC achieved by the ablation model without $GCN_{D}$ exceeded that of the entire model by only 0.01\%, which is within an acceptable fluctuation range. For the other datasets, the gap between the entire model and the one without $GCN_{X}$ is remarkable, which manifests that node features play an essential role in our model. Furthermore, the model's performance is markedly diminished when the loss corresponding to a specific class is eliminated, suggesting that the model requires losses from both positive and negative samples to retain efficacy, and models with only one loss can not capture the full information. Notably, the loss associated with normal graphs appears to exert a more substantial influence on the model's ability. All the results state that IGAD-CF is able to effectively enhance recognition performance after implementing sample balancing and assigning weights to features. And the combination of degree attributes and node features brings the greatest progress to the model.

\subsection{Available Feature Analysis}

Due to incomplete node features in some datasets, when we apply the methods to the datasets mentioned, we disregard the node attributes provided in the original datasets and instead derive node features solely from the graph adjacency matrices. For each adjacency matrix $\mathbf{A}_i$ from a graph, we explore three different approaches to construct the corresponding node feature matrix \cite{cui2022positional}.

\emph{Identity Encoding:} Each node is transferred by a one-hot encoding vector. Formally, a node feature matrix $\mathbf{X}_i$ is constructed by repeating the identity matrix $\mathbf{I}_i$ $N$ times along the diagonal, where $N$ represents the nodes' number in the graph.

\emph{Degree Binning (DB):} After calculating each node's degree, these degrees $\mathbf{D}$ are discretized into predefined bins. During the processing, a histogram where each bin corresponds to a range of degrees is created. Then we encode the nodes' degrees to put them into the bins respectively.

\emph{Local Degree Profile (LDP):} This approach uses both the individual node's degree and statistical properties of the degrees of the neighbors to construct node features.

\subsubsection{Performance Comparison}
By exploring these three distinct methods for constructing node features from the adjacency matrices, our aim is to examine the impact of varying node features and their impact on the performance of IGAD-CF. The results are displayed in Table \ref{feature table1}.
For most cases, the identity method outperforms others, with the LDP method generally yielding the least favorable results. This suggests that the identity method is the most effective in highlighting the distinctive attributes of nodes.

\subsubsection{Validity Verification}

To provide validation for the efficacy of our feature construction approach across various datasets, we also conduct comparative experiments with the baseline methods GLocalKD, GOOD-D, and GLADST on six datasets. Specifically, we train the baseline models on the selected datasets using the initial node features from the datasets or the node features constructed via Identity Encoding separately. Subsequently, we evaluated the trained models and recorded the corresponding results. As shown in Table \ref{feature table2}, except for the ENZYMES dataset where our constructed node features performed slightly weaker than the original node features, our method exhibited significantly superior performance on the other datasets.

\begin{figure*}
\centering
\subfigure[AIDS on IGAD-CF]{
\includegraphics[width=0.3\textwidth]{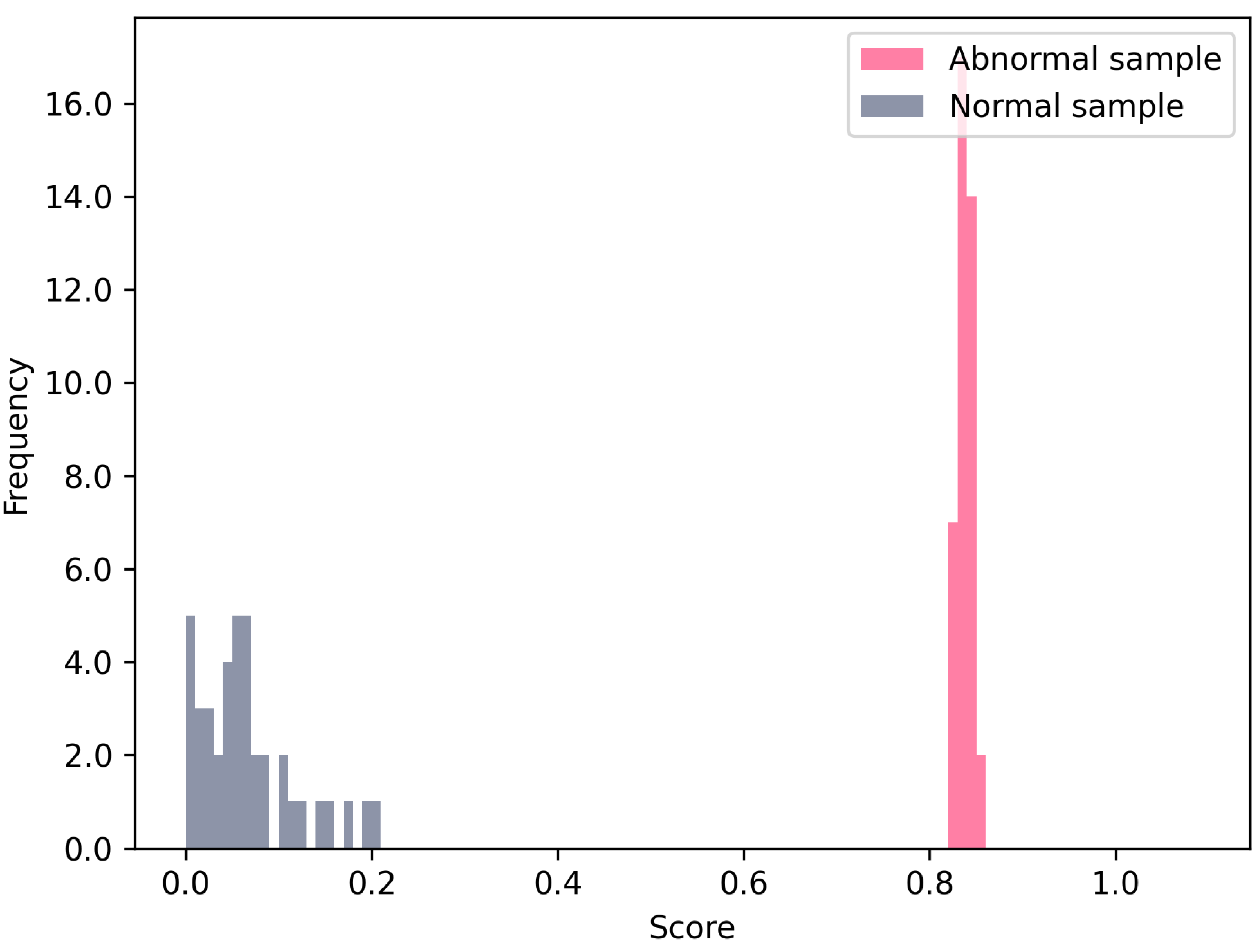}
\label{fig:IGAD-CF_AIDS}}
\subfigure[DHFR on IGAD-CF]{
\includegraphics[width=0.3\textwidth]{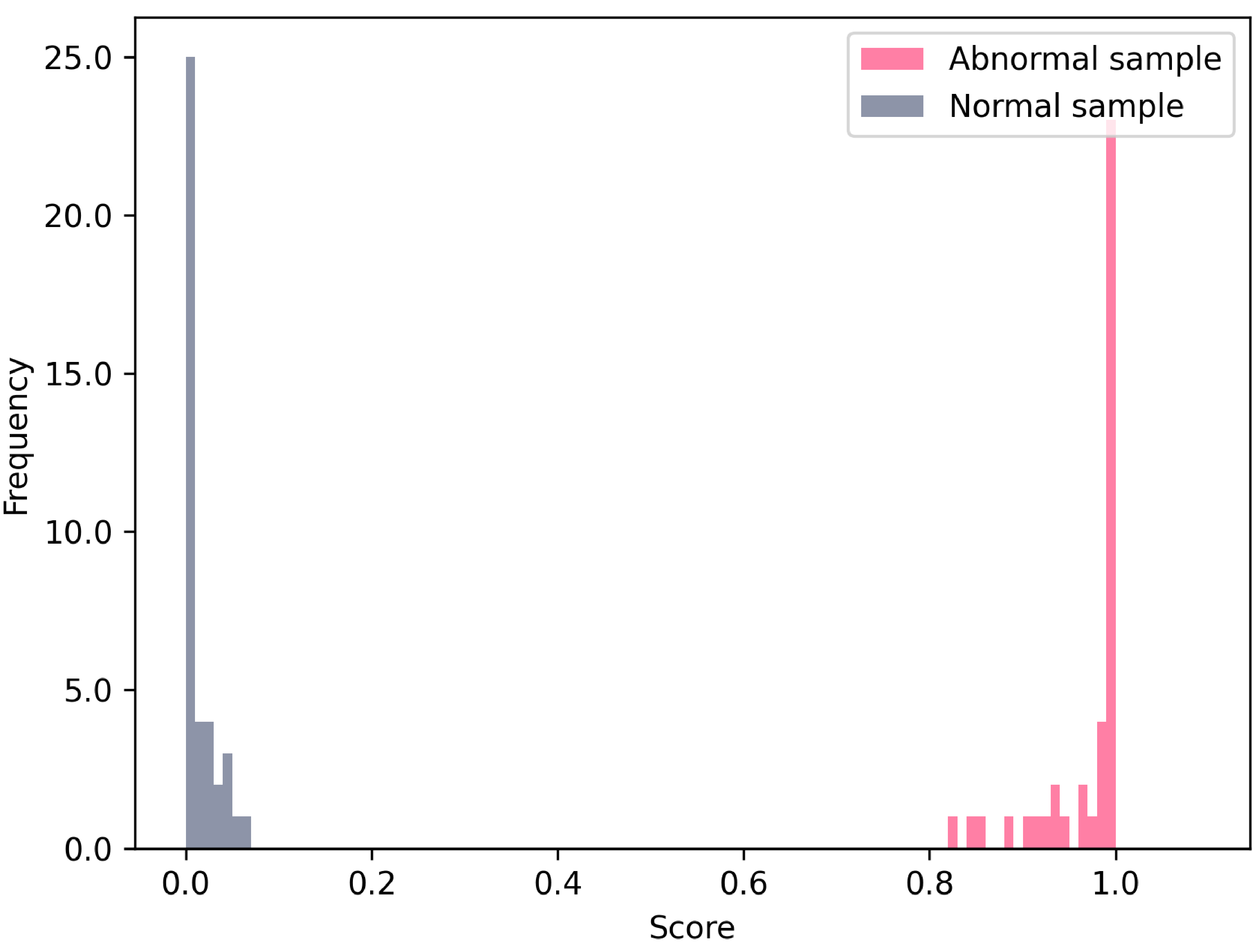} 
\label{fig:IGAD-CF_DHFR}}
\subfigure[NCI1 on IGAD-CF]{
\includegraphics[width=0.3\textwidth]{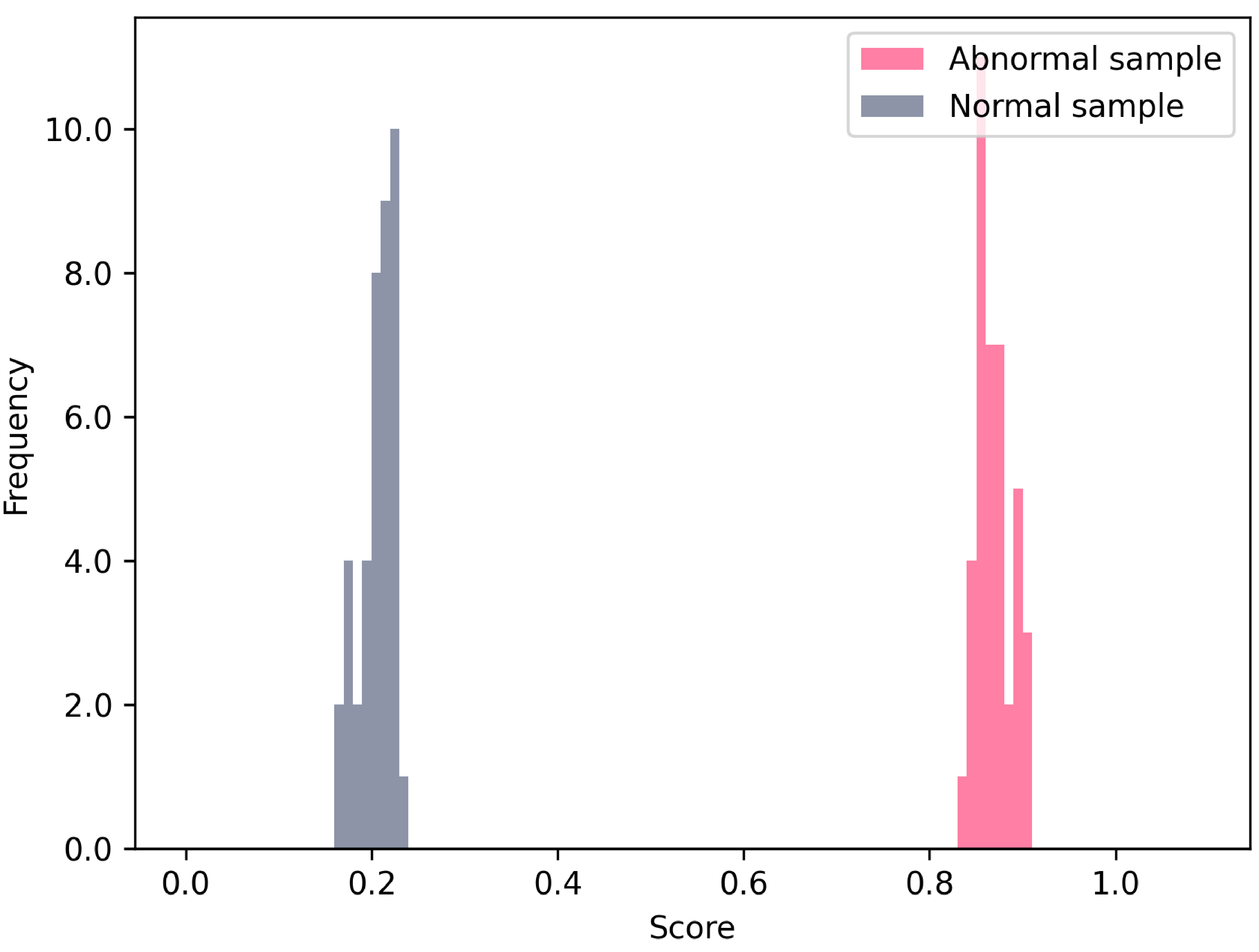} 
\label{fig:IGAD-CF_NCI1}}

\subfigure[AIDS on GLocalKD]{
\includegraphics[width=0.3\textwidth]{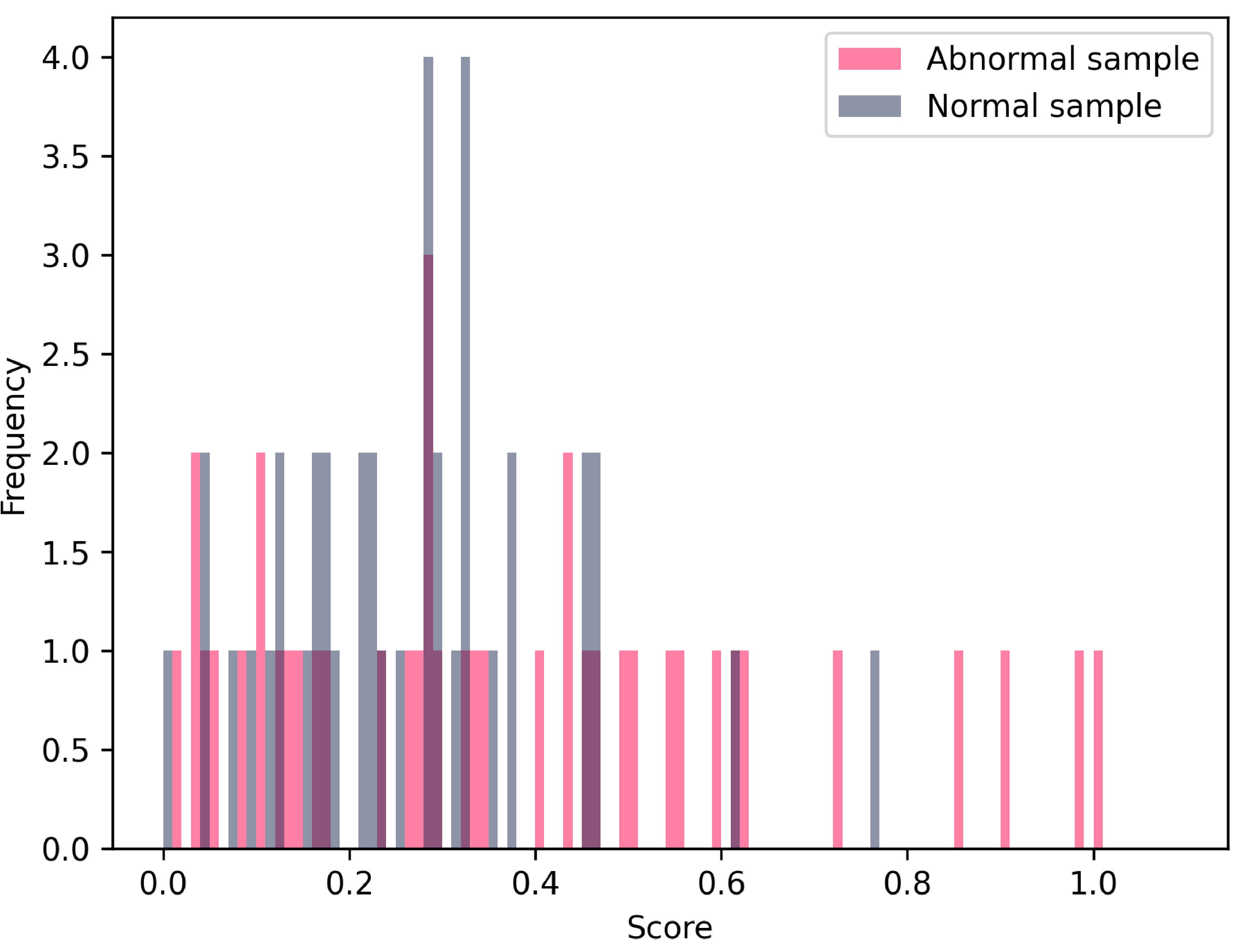}
\label{fig:GLocalKD_AIDS}}
\subfigure[DHFR on GLocalKD]{
\includegraphics[width=0.3\textwidth]{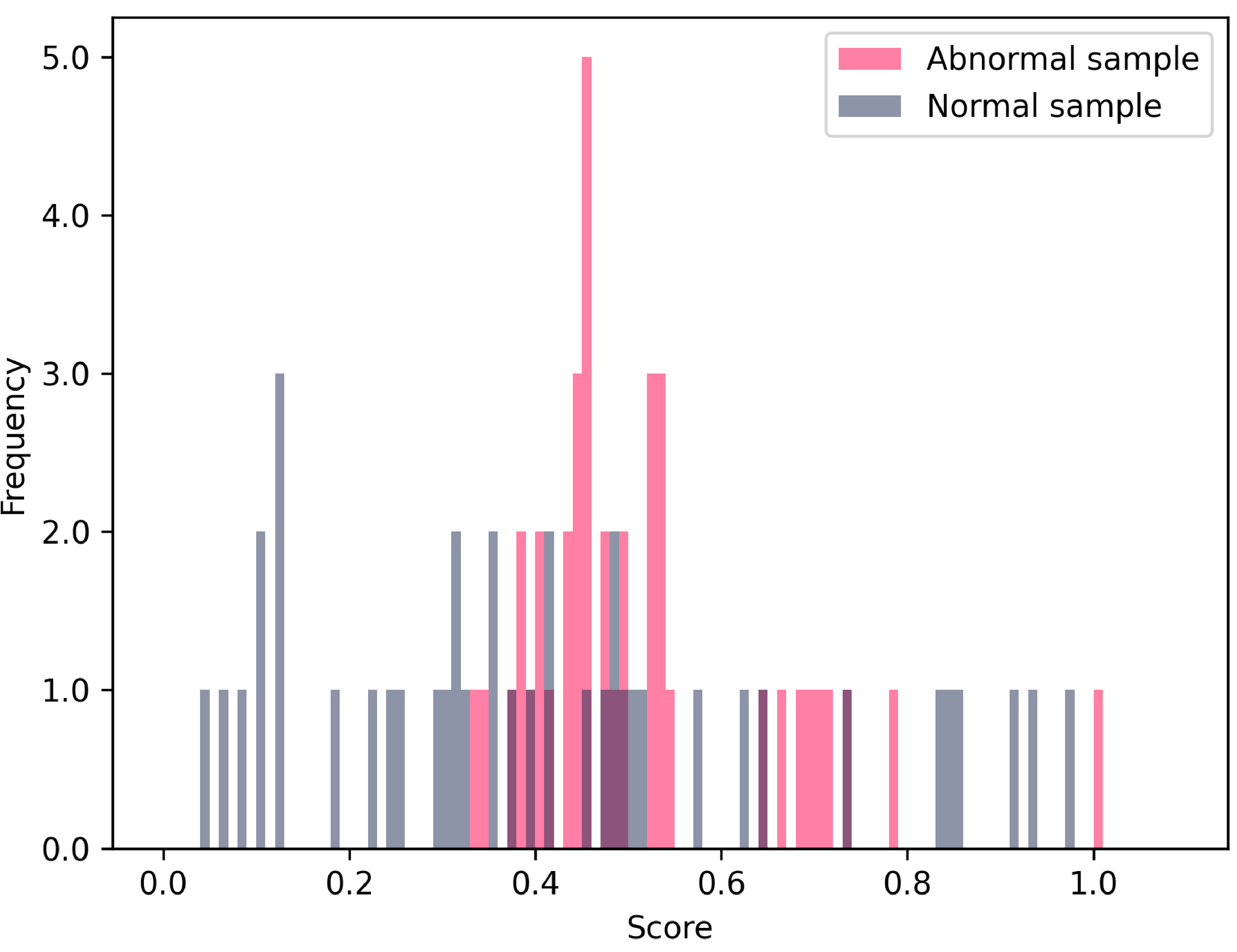} 
\label{fig:GLocalKD_DHFR}}
\subfigure[NCI1 on GLocalKD]{
\includegraphics[width=0.3\textwidth]{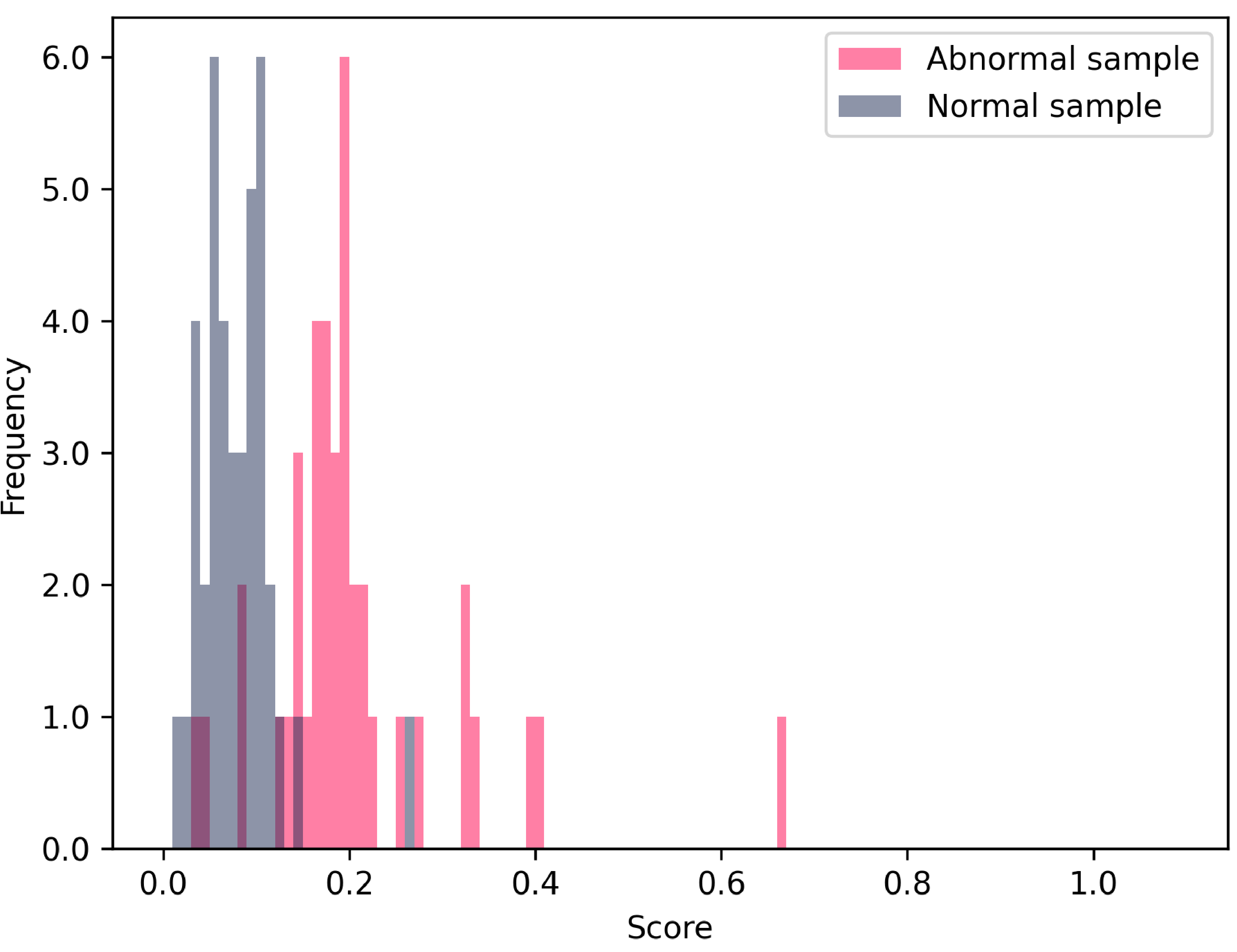} 
\label{fig:GLocalKD_NCI1}}

\subfigure[AIDS on GLADST]{
\includegraphics[width=0.3\textwidth]{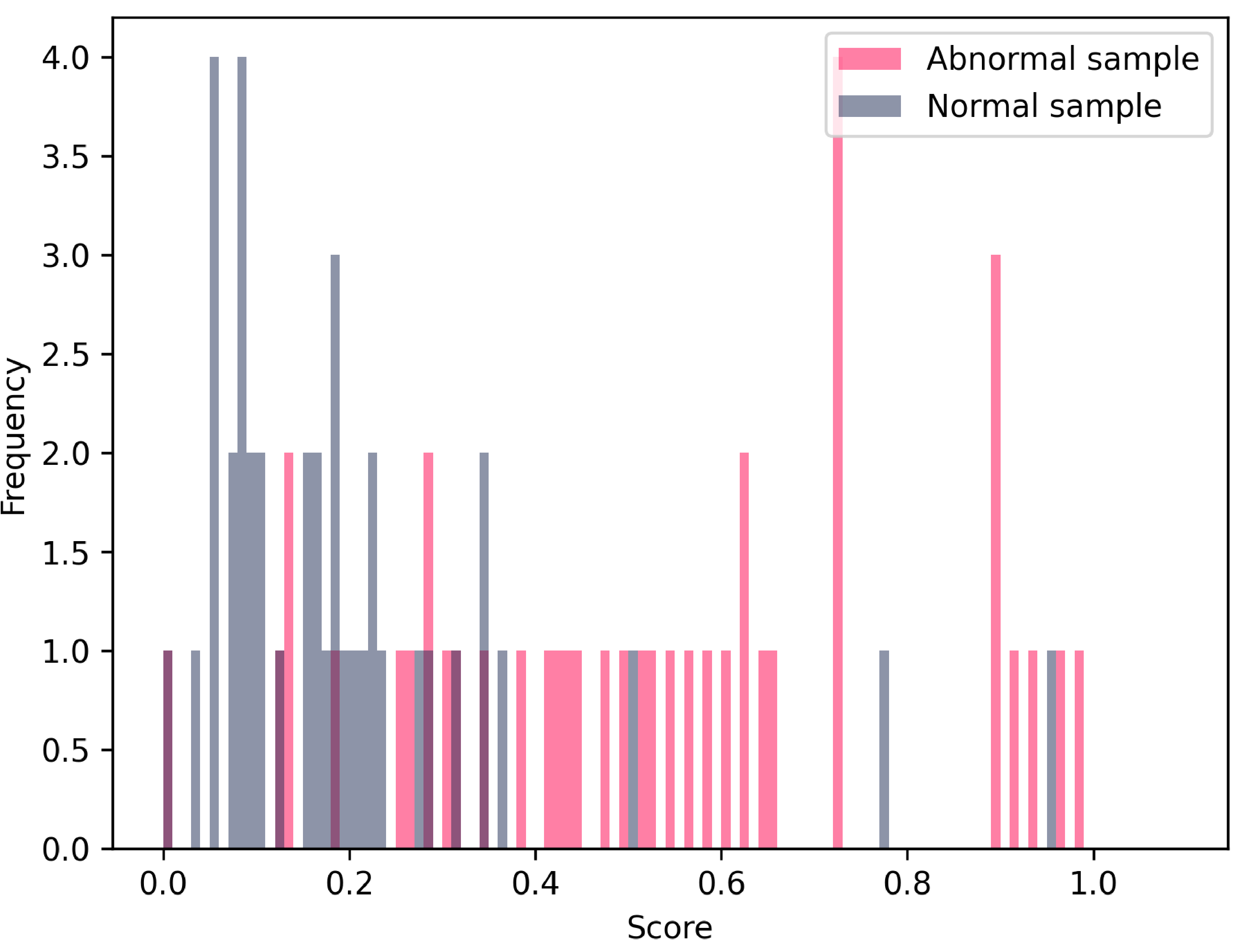}
\label{fig:GLADST_AIDS}}
\subfigure[DHFR on GLADST]{
\includegraphics[width=0.3\textwidth]{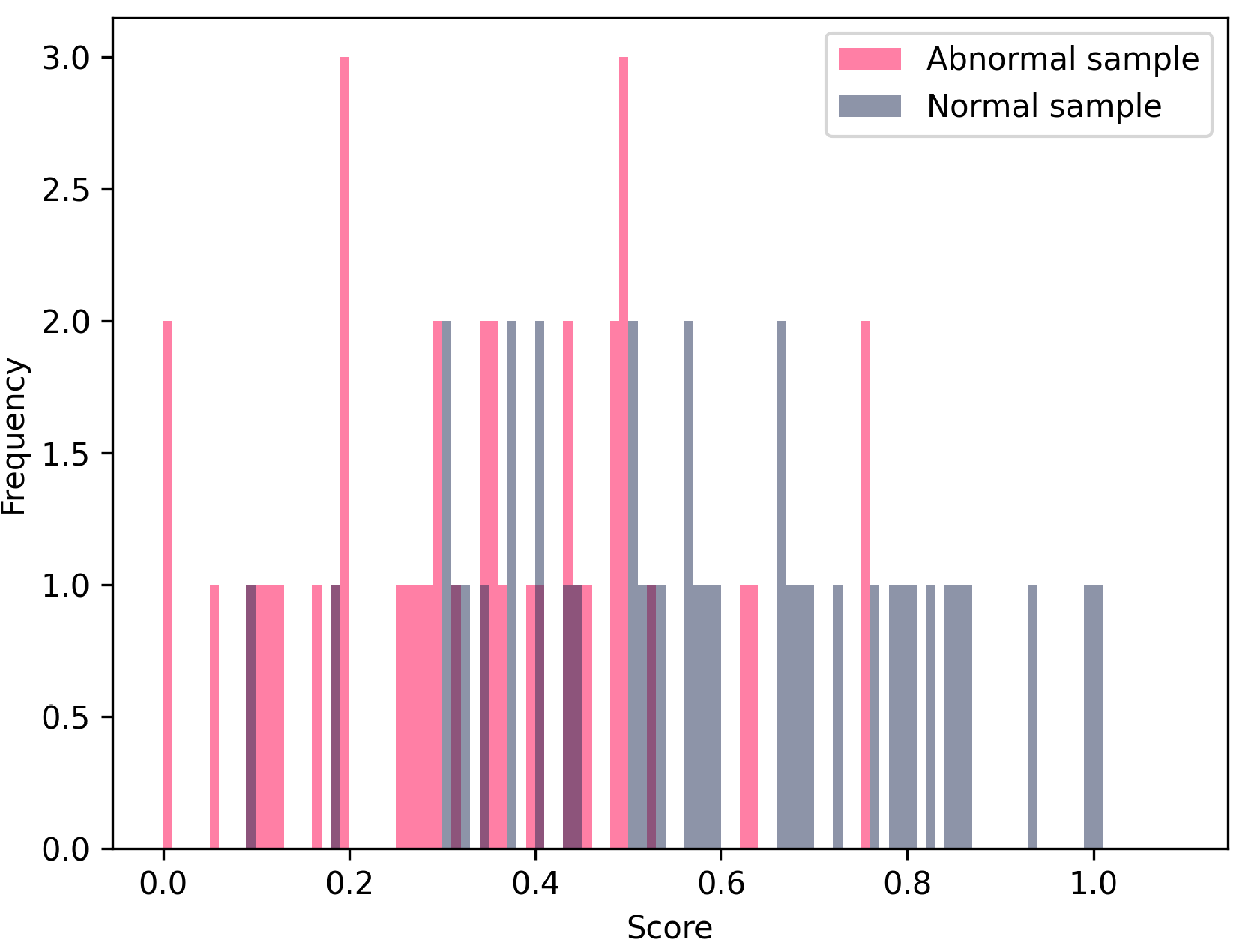} 
\label{fig:GLADST_DHFR}}
\subfigure[NCI1 on GLADST]{
\includegraphics[width=0.3\textwidth]{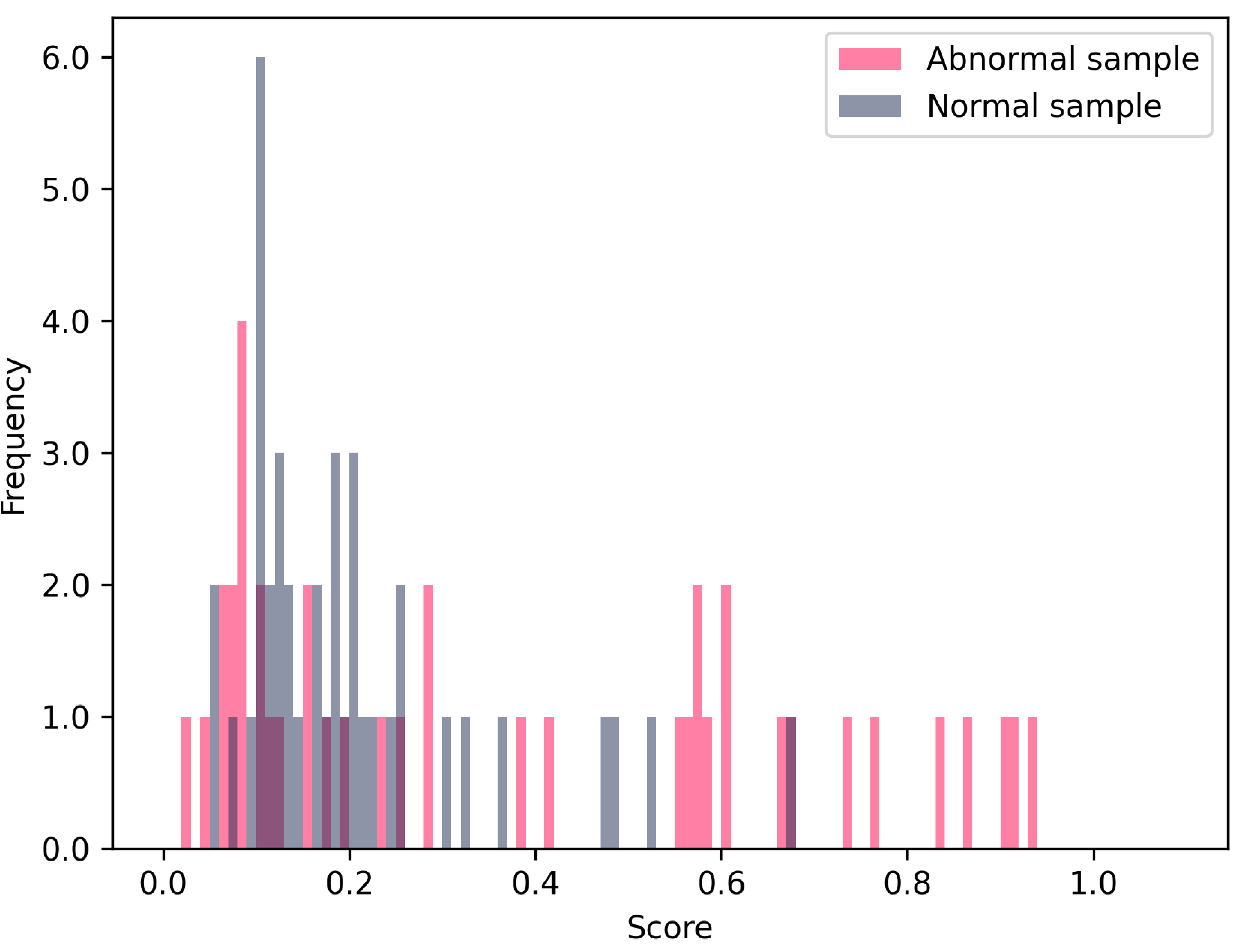} 
\label{fig:GLADST_NCI1}}

\caption{A visualization analysis for IGAD-CF's and selected baselines' performance on the selected datasets, where the pink and grey represent abnormal samples and normal samples. When the pink is concentrated on the right side of the axis, and the grey is concentrated on the left side of the axis, it indicates that the model is better can classify the samples better.}
\label{fig:Visual}
\end{figure*}

\subsection{Visualization Analysis}

To visually reflect the classification performance of our model on normal and abnormal samples from a more intuitive perspective, we visualized the anomaly scores of the node identification test set samples through bar charts.
Here we choose AIDS, DHFR, and NCI1 for verification. GLADST and GLocalKD are selected to compare with our model. The results are shown in Figure \ref{fig:Visual}.

The abscissa of these histograms represents the anomaly scores, and the ordinate is the number of corresponding samples with anomaly scores falling within that range. An anomaly score that is closer to 1 means a higher likelihood that this object is an abnormal sample, while a score closer to 0 suggests that the sample is normal. these histograms reveal that, for the majority of cases, the abnormal and normal samples tend to cluster on opposite ends of the axis. This demonstrates that IGAD-CF is capable of distinguishing normal and anomalous samples to a significant extent, while the two baselines selected are not ideal. It can be seen clearly in the figures that the baselines failed to widen the gap between the normal and the abnormal, which leads to poor classification effects.

\subsection{Parameter Analysis}
As outlined in the method section, we introduced a hyperparameter, $\beta$, to manually control the impact of our generated abnormal samples in the framework's final decision-making process. In this section, we will focus on investigating its impact on model performance, and observe whether the optimal $\beta$ changes in different datasets. Thus, we have selected four datasets (AIDS, BZR, COX2, DHFR, and ENZYMES) for our experimental evaluation.

\begin{figure}[!h]
\centering
\subfigure[AIDS]{
\includegraphics[width=0.2\textwidth]{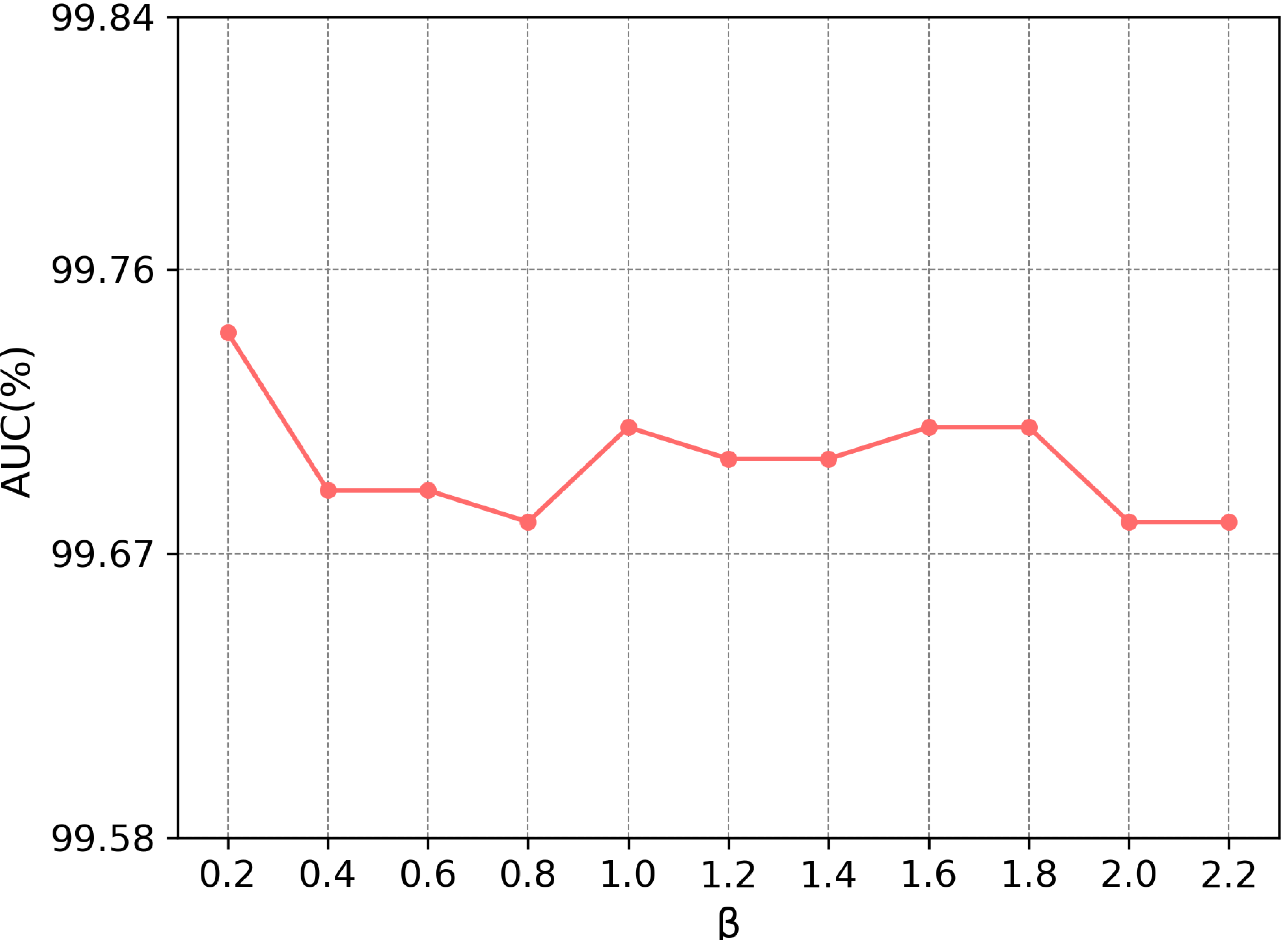}
\label{fig:AIDSP}}
\subfigure[BZR]{
\includegraphics[width=0.2\textwidth]{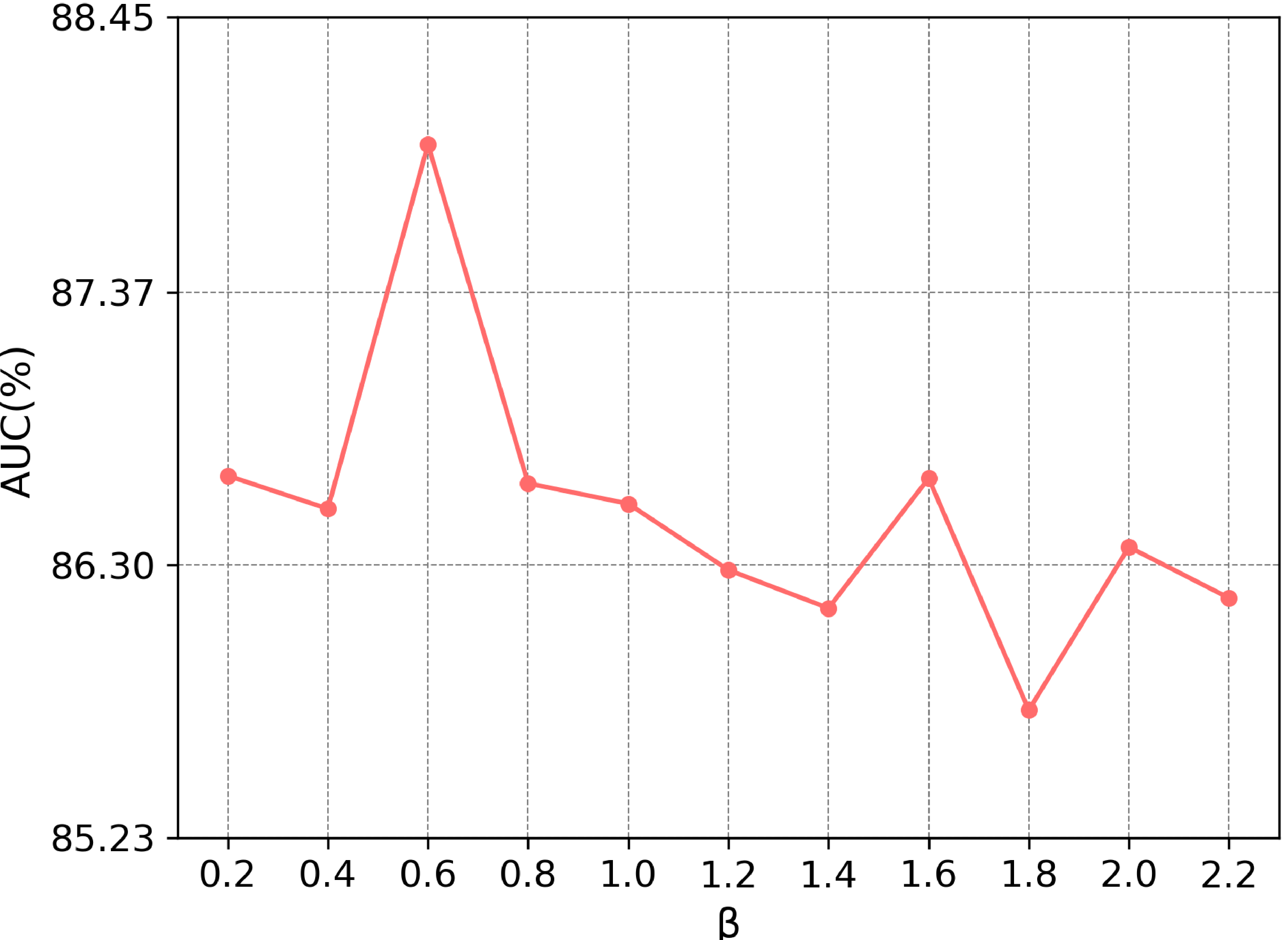}
\label{fig:BZRP}}
\subfigure[COX2]{
\includegraphics[width=0.2\textwidth]{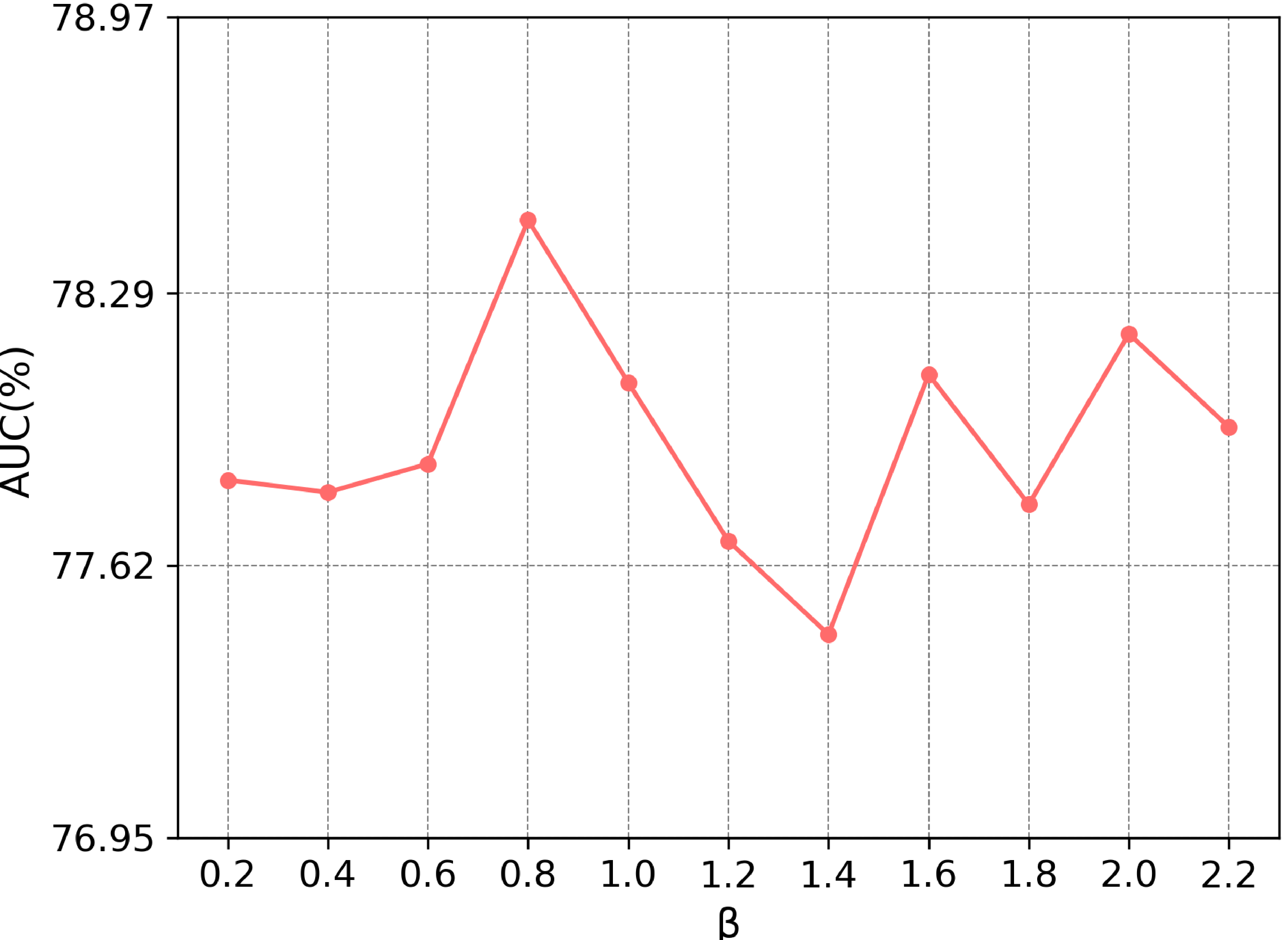} 
\label{fig:COX2P}}
\subfigure[DHFR]{
\includegraphics[width=0.2\textwidth]{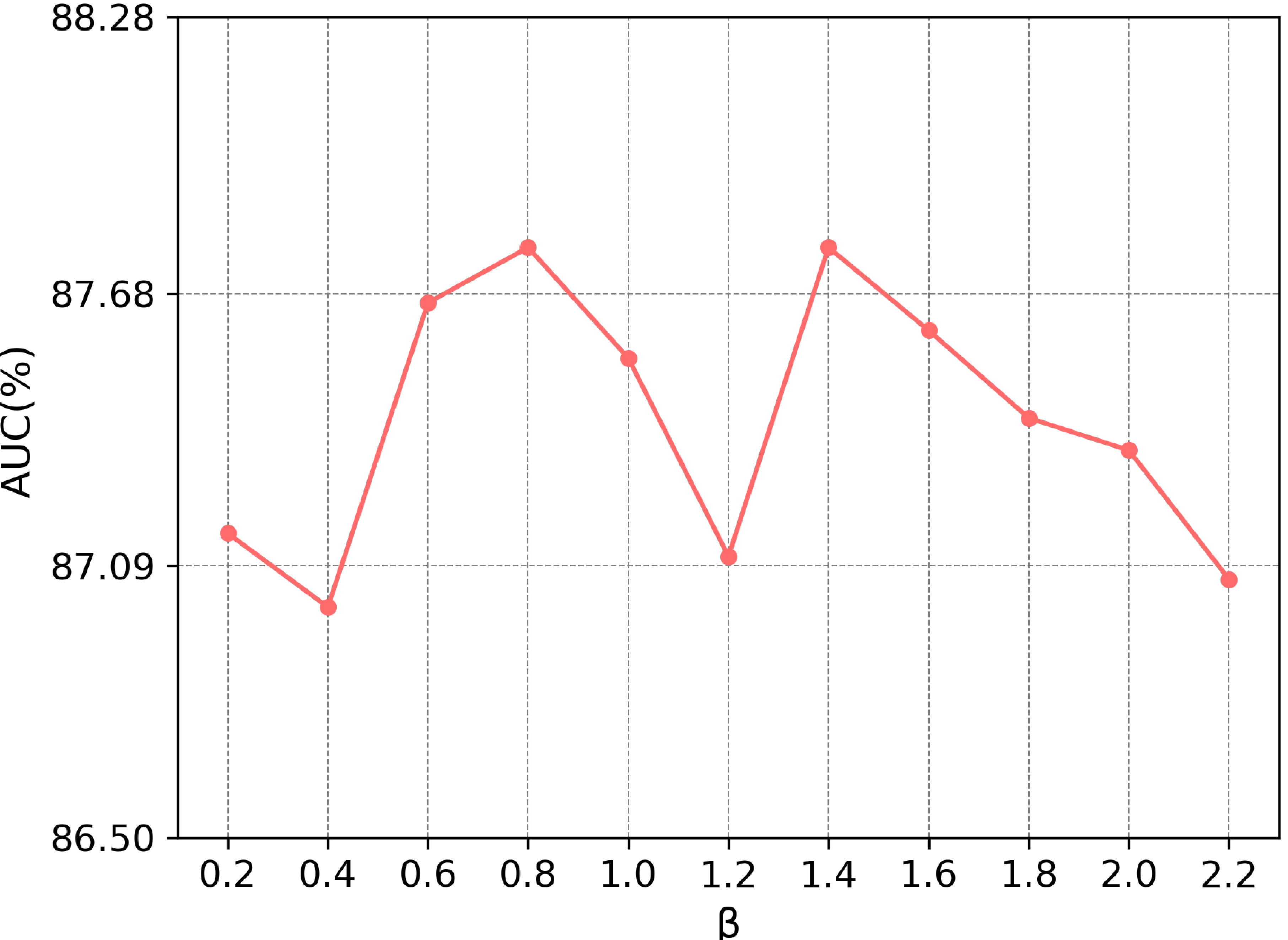} 
\label{fig:DHFRP}}
\subfigure[ENZYMES]{
\includegraphics[width=0.2\textwidth]{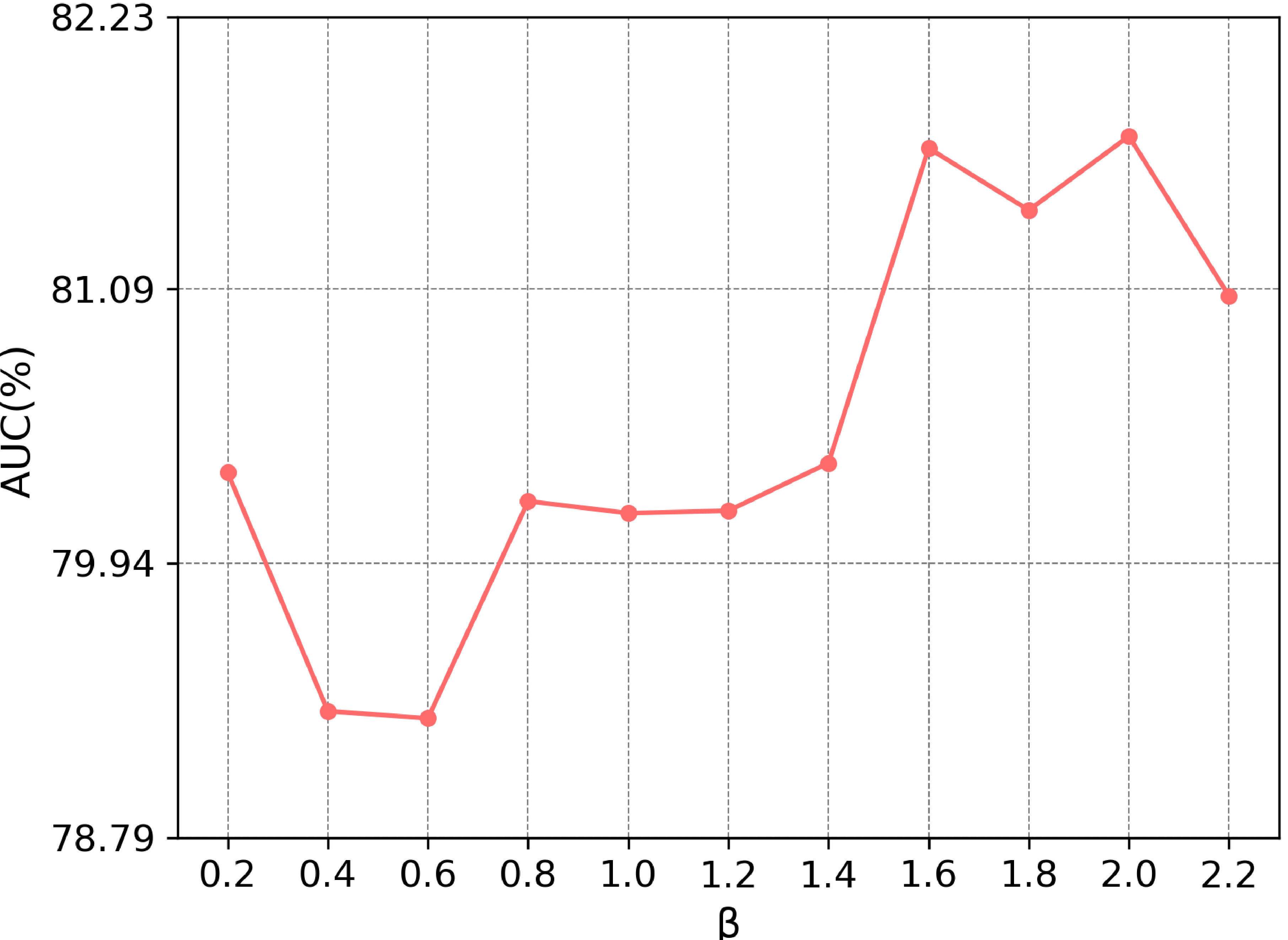} 
\label{fig:ENZYMESP}}
\subfigure[hERG]{
\includegraphics[width=0.2\textwidth]{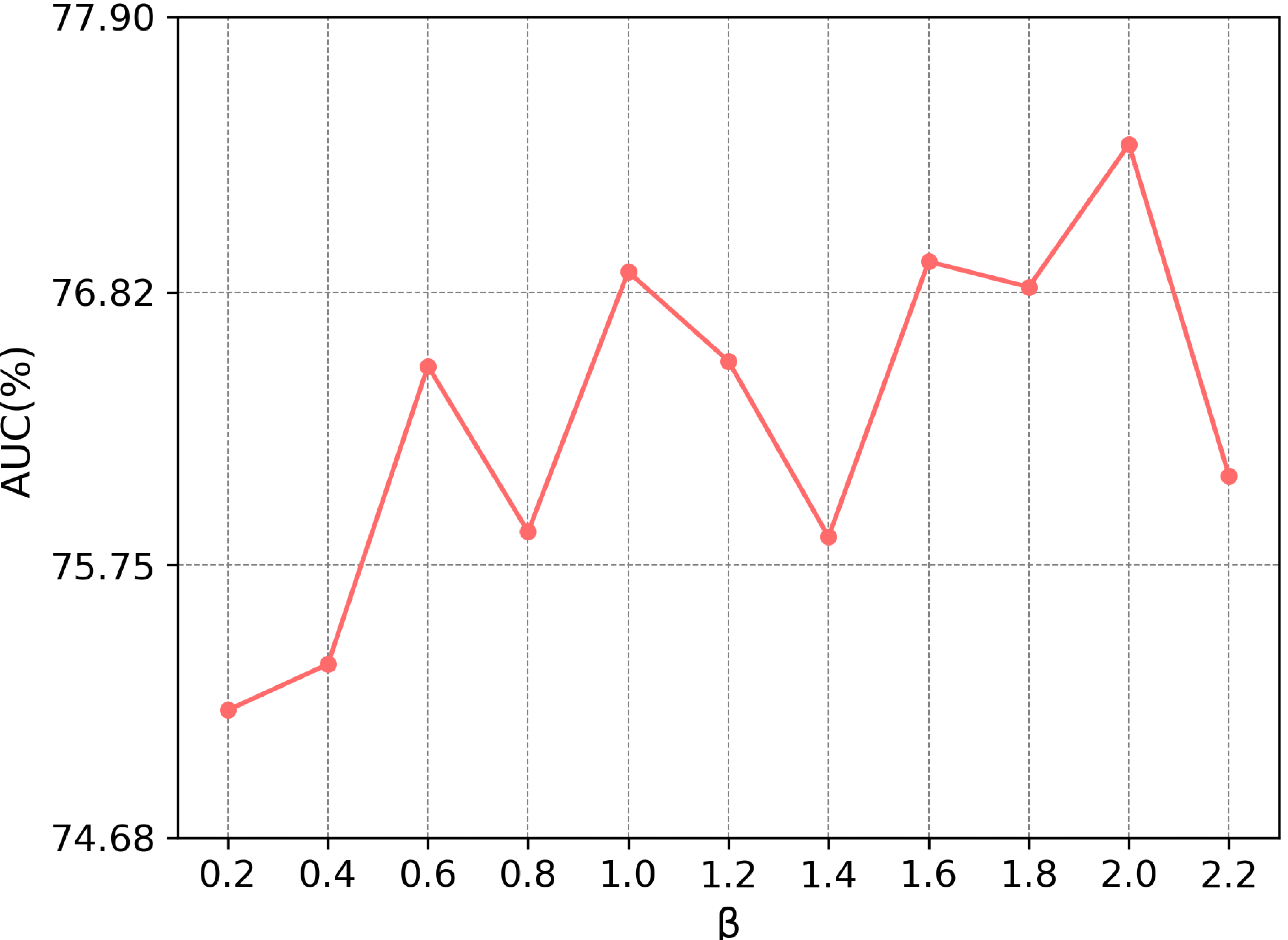} 
\label{fig:hERGP}}

\caption{Parametric experiments on hyperparametric $\beta$. It represents the weights of our expanded samples in participating during the overall training.}
\label{fig: Parameter}
\end{figure}

In Figure \ref{fig: Parameter}, the abscissa represents the values of $\beta$, while the ordinate corresponds to the AUC values. The graph illustrates that variations in beta between 0.2 and 2.2 have a slight impact on the model’s performance. Notably, when $\beta$ is set between 1.2 and 1.4 or 
 between 0.6 and 0.8, the overall effect is more favorable.

\subsection{Case Study for Disorder Detection}
In addition to achieving satisfactory results on usual graph anomaly detection datasets, we extend the downstream applications of IGAD-CF by applying it to the identification of abnormal brain graphs for brain disease detection. This extension serves to validate the robust generalization capabilities and practical utility of IGAD-CF. 
\subsubsection{Datasets from Brain Graphs}
We select two brain disorder datasets, human immunodeficiency virus infection (HIV) and bipolar disorder (BP), each encompassing two neuroimaging modalities functional diffusion tensor imaging (DTI) and magnetic resonance imaging (fMRI) \cite{cui2022interpretable}. Consequently, four datasets are formed: HIV-DTI, HIV-fMRI, BP-DTI, and BP-fMRI. We construct the brain graphs by employing previously established techniques from brain network studies \cite{cui2022interpretable,cui2022braingb}. Specifically, for each subject in the datasets, the neuroimaging data is used to construct a brain graph. The nodes from the constructed graph represent distinct brain regions, and the edges represent the structural or functional relationships between these regions. Each HIV dataset comprises a total of 70 samples, evenly distributed between normal and anomalous cases. For every graph in the HIV dataset, there are 90 valid brain regions, leading to the construction of 90 nodes. Each BP dataset contains 97 samples in total, with 52 representing anomalous brain graphs and the remaining 45 being normal cases. Each sample in the BP dataset consists of 82 nodes.
\begin{table}[!h]
\centering
\caption{The mean value of AUC (\%) and standard deviation. This table reports the results obtained by two widely used brain graph anomaly detection methods and our proposed IGAD-CF framework when trained on four baseline datasets.}
\begin{tabular}{c|ccc}
\hline
Datasets  & BrainGNN \cite{li2021braingnn}    & IBGNN \cite{cui2022interpretable}       & IGAD-CF               \\ \hline
HIV-DTI  & 68.91\tiny±7.07  & 51.01\tiny±10.53 & \textbf{77.55\tiny±7.30}  \\
HIV-fMRI & 68.10\tiny±9.62  & 59.90\tiny±8.24  & \textbf{73.06\tiny±8.40}  \\
BP-DTI   & 57.45\tiny±8.31  & 52.05\tiny±5.73  & \textbf{59.31\tiny±12.88} \\
BP-fMRI  & 55.71\tiny±12.97 & 51.37\tiny±8.03  & \textbf{73.11\tiny±14.39} \\ \hline
\end{tabular}
\label{Brain table1}
\end{table}
\subsubsection{Baselines}
To demonstrate the efficacy of our approach in brain graph experiments, we select two deep learning baselines for comparison. These two methods are popular deep learning methods in brain graph detection, which can capture brain graph anomalies well in the application field.
BrainGNN \cite{li2021braingnn} designs a selection pooling layer, which is capable of focusing on the most representative brain regions automatically. IBGNN \cite{cui2022interpretable} consists of a message-passing-based backbone and an explanation generator based on a global-shared mask matrix.

\subsubsection{Performance Analysis}
We conducted experiments on the brain graph datasets. Similar to previous studies on classic anomaly detection datasets, we record the mean AUC and standard deviation for evaluation. The final results are presented in Table \ref{Brain table1}, which showcases that IGAD-CF significantly outstrips the two baselines across all datasets. Moreover, our method demonstrates markedly superior stability compared to the other methods in the HIV datasets. The satisfying result further improves the feasibility of applying our approach to the detection of brain diseases.

\section{Conclusion}
In our proposed work, to address the key challenges in the GLAD task, we design the IGAD-CF framework. To resolve the performance degradation caused by sample imbalance, we construct an anomaly sample generation module that leverages counterfactual reasoning mechanisms to generate anomalous samples, thereby enriching the datasets. Furthermore, to address the issue of existing methods not fully utilizing degree attributes and node features, our node feature learning module innovatively integrates the two, allowing for complementary information exchange and presenting more comprehensive information. Subsequently, we employ an adaptive weight matrix to emphasize the extracted graph features at different levels of importance. Finally, we carry out a comprehensive set of experiments that not only demonstrate our model's outstanding performance on traditional public datasets but also exhibit its significant potential in brain graph anomaly detection.

\begin{acks}
We would like to thank the anonymous reviewers for their helpful comments and feedback. This research was supported by Wuhan University People's Hospital Cross-Innovation Talent Project Foundation under JCRCZN-2022-008. 
\end{acks}

\balance
\bibliographystyle{ACM-Reference-Format}
\bibliography{ref}


\begin{thebibliography}{48}


\ifx \showCODEN    \undefined \def \showCODEN     #1{\unskip}     \fi
\ifx \showDOI      \undefined \def \showDOI       #1{#1}\fi
\ifx \showISBNx    \undefined \def \showISBNx     #1{\unskip}     \fi
\ifx \showISBNxiii \undefined \def \showISBNxiii  #1{\unskip}     \fi
\ifx \showISSN     \undefined \def \showISSN      #1{\unskip}     \fi
\ifx \showLCCN     \undefined \def \showLCCN      #1{\unskip}     \fi
\ifx \shownote     \undefined \def \shownote      #1{#1}          \fi
\ifx \showarticletitle \undefined \def \showarticletitle #1{#1}   \fi
\ifx \showURL      \undefined \def \showURL       {\relax}        \fi
\providecommand\bibfield[2]{#2}
\providecommand\bibinfo[2]{#2}
\providecommand\natexlab[1]{#1}
\providecommand\showeprint[2][]{arXiv:#2}

\bibitem[Breunig et~al\mbox{.}(2000)]%
        {breunig2000lof}
\bibfield{author}{\bibinfo{person}{Markus~M Breunig}, \bibinfo{person}{Hans-Peter Kriegel}, \bibinfo{person}{Raymond~T Ng}, {and} \bibinfo{person}{J{\"o}rg Sander}.} \bibinfo{year}{2000}\natexlab{}.
\newblock \showarticletitle{LOF: identifying density-based local outliers}. In \bibinfo{booktitle}{\emph{SIGMOD}}. \bibinfo{pages}{93--104}.
\newblock


\bibitem[Byrne(2019)]%
        {byrne2019counterfactuals}
\bibfield{author}{\bibinfo{person}{Ruth~MJ Byrne}.} \bibinfo{year}{2019}\natexlab{}.
\newblock \showarticletitle{Counterfactuals in Explainable Artificial Intelligence (XAI): Evidence from Human Reasoning.}. In \bibinfo{booktitle}{\emph{IJCAI}}. \bibinfo{pages}{6276--6282}.
\newblock


\bibitem[Chaudhary et~al\mbox{.}(2019)]%
        {8862186}
\bibfield{author}{\bibinfo{person}{Anshika Chaudhary}, \bibinfo{person}{Himangi Mittal}, {and} \bibinfo{person}{Anuja Arora}.} \bibinfo{year}{2019}\natexlab{}.
\newblock \showarticletitle{Anomaly Detection using Graph Neural Networks}. In \bibinfo{booktitle}{\emph{COMITCon}}. \bibinfo{pages}{346--350}.
\newblock
\urldef\tempurl%
\url{https://doi.org/10.1109/COMITCon.2019.8862186}
\showDOI{\tempurl}


\bibitem[Chen et~al\mbox{.}(2023)]%
        {chen2023counterfactual}
\bibfield{author}{\bibinfo{person}{Long Chen}, \bibinfo{person}{Yuhang Zheng}, \bibinfo{person}{Yulei Niu}, \bibinfo{person}{Hanwang Zhang}, {and} \bibinfo{person}{Jun Xiao}.} \bibinfo{year}{2023}\natexlab{}.
\newblock \showarticletitle{Counterfactual samples synthesizing and training for robust visual question answering}.
\newblock \bibinfo{journal}{\emph{IEEE Transactions on Pattern Analysis and Machine Intelligence}} (\bibinfo{year}{2023}).
\newblock


\bibitem[Cui et~al\mbox{.}(2022a)]%
        {cui2022braingb}
\bibfield{author}{\bibinfo{person}{Hejie Cui}, \bibinfo{person}{Wei Dai}, \bibinfo{person}{Yanqiao Zhu}, \bibinfo{person}{Xuan Kan}, \bibinfo{person}{Antonio Aodong~Chen Gu}, \bibinfo{person}{Joshua Lukemire}, \bibinfo{person}{Liang Zhan}, \bibinfo{person}{Lifang He}, \bibinfo{person}{Ying Guo}, {and} \bibinfo{person}{Carl Yang}.} \bibinfo{year}{2022}\natexlab{a}.
\newblock \showarticletitle{Braingb: a benchmark for brain network analysis with graph neural networks}.
\newblock \bibinfo{journal}{\emph{IEEE Transactions on Medical Imaging}} \bibinfo{volume}{42}, \bibinfo{number}{2} (\bibinfo{year}{2022}), \bibinfo{pages}{493--506}.
\newblock


\bibitem[Cui et~al\mbox{.}(2022b)]%
        {cui2022interpretable}
\bibfield{author}{\bibinfo{person}{Hejie Cui}, \bibinfo{person}{Wei Dai}, \bibinfo{person}{Yanqiao Zhu}, \bibinfo{person}{Xiaoxiao Li}, \bibinfo{person}{Lifang He}, {and} \bibinfo{person}{Carl Yang}.} \bibinfo{year}{2022}\natexlab{b}.
\newblock \showarticletitle{Interpretable graph neural networks for connectome-based brain disorder analysis}. In \bibinfo{booktitle}{\emph{MICCAI}}. \bibinfo{pages}{375--385}.
\newblock


\bibitem[Cui et~al\mbox{.}(2022c)]%
        {cui2022positional}
\bibfield{author}{\bibinfo{person}{Hejie Cui}, \bibinfo{person}{Zijie Lu}, \bibinfo{person}{Pan Li}, {and} \bibinfo{person}{Carl Yang}.} \bibinfo{year}{2022}\natexlab{c}.
\newblock \showarticletitle{On positional and structural node features for graph neural networks on non-attributed graphs}. In \bibinfo{booktitle}{\emph{CIKM}}. \bibinfo{pages}{3898--3902}.
\newblock


\bibitem[Fern{\'a}ndez et~al\mbox{.}(2019)]%
        {fernandez2019relevance}
\bibfield{author}{\bibinfo{person}{Rub{\'e}n~R Fern{\'a}ndez}, \bibinfo{person}{Isaac~Mart{\'\i}n de Diego}, \bibinfo{person}{V{\'\i}ctor Ace{\~n}a}, \bibinfo{person}{Javier~M Moguerza}, {and} \bibinfo{person}{Alberto Fern{\'a}ndez-Isabel}.} \bibinfo{year}{2019}\natexlab{}.
\newblock \showarticletitle{Relevance metric for counterfactuals selection in decision trees}. In \bibinfo{booktitle}{\emph{IDEAL}}. \bibinfo{pages}{85--93}.
\newblock


\bibitem[Gori et~al\mbox{.}(2005)]%
        {gori2005new}
\bibfield{author}{\bibinfo{person}{Marco Gori}, \bibinfo{person}{Gabriele Monfardini}, {and} \bibinfo{person}{Franco Scarselli}.} \bibinfo{year}{2005}\natexlab{}.
\newblock \showarticletitle{A new model for learning in graph domains}. In \bibinfo{booktitle}{\emph{IJCNN}}, Vol.~\bibinfo{volume}{2}. \bibinfo{pages}{729--734}.
\newblock


\bibitem[Hendricks et~al\mbox{.}(2018)]%
        {hendricks2018grounding}
\bibfield{author}{\bibinfo{person}{Lisa~Anne Hendricks}, \bibinfo{person}{Ronghang Hu}, \bibinfo{person}{Trevor Darrell}, {and} \bibinfo{person}{Zeynep Akata}.} \bibinfo{year}{2018}\natexlab{}.
\newblock \showarticletitle{Grounding visual explanations}. In \bibinfo{booktitle}{\emph{ECCV}}. \bibinfo{pages}{264--279}.
\newblock


\bibitem[Huang et~al\mbox{.}(2018)]%
        {8325544}
\bibfield{author}{\bibinfo{person}{Dongxu Huang}, \bibinfo{person}{Dejun Mu}, \bibinfo{person}{Libin Yang}, {and} \bibinfo{person}{Xiaoyan Cai}.} \bibinfo{year}{2018}\natexlab{}.
\newblock \showarticletitle{CoDetect: Financial Fraud Detection With Anomaly Feature Detection}.
\newblock \bibinfo{journal}{\emph{IEEE Access}}  \bibinfo{volume}{6} (\bibinfo{year}{2018}), \bibinfo{pages}{19161--19174}.
\newblock


\bibitem[Huang et~al\mbox{.}(2022)]%
        {huang2022dgraph}
\bibfield{author}{\bibinfo{person}{Xuanwen Huang}, \bibinfo{person}{Yang Yang}, \bibinfo{person}{Yang Wang}, \bibinfo{person}{Chunping Wang}, \bibinfo{person}{Zhisheng Zhang}, \bibinfo{person}{Jiarong Xu}, \bibinfo{person}{Lei Chen}, {and} \bibinfo{person}{Michalis Vazirgiannis}.} \bibinfo{year}{2022}\natexlab{}.
\newblock \showarticletitle{Dgraph: A large-scale financial dataset for graph anomaly detection}.
\newblock \bibinfo{journal}{\emph{NeurIPS}}  \bibinfo{volume}{35} (\bibinfo{year}{2022}), \bibinfo{pages}{22765--22777}.
\newblock


\bibitem[Jiang et~al\mbox{.}(2019)]%
        {9020760}
\bibfield{author}{\bibinfo{person}{Jianguo Jiang}, \bibinfo{person}{Jiuming Chen}, \bibinfo{person}{Tianbo Gu}, \bibinfo{person}{Kim-Kwang~Raymond Choo}, \bibinfo{person}{Chao Liu}, \bibinfo{person}{Min Yu}, \bibinfo{person}{Weiqing Huang}, {and} \bibinfo{person}{Prasant Mohapatra}.} \bibinfo{year}{2019}\natexlab{}.
\newblock \showarticletitle{Anomaly Detection with Graph Convolutional Networks for Insider Threat and Fraud Detection}. In \bibinfo{booktitle}{\emph{MILCOM}}. \bibinfo{pages}{109--114}.
\newblock
\urldef\tempurl%
\url{https://doi.org/10.1109/MILCOM47813.2019.9020760}
\showDOI{\tempurl}


\bibitem[Kahneman and Tversky(1981)]%
        {kahneman1981simulation}
\bibfield{author}{\bibinfo{person}{Daniel Kahneman} {and} \bibinfo{person}{Amos Tversky}.} \bibinfo{year}{1981}\natexlab{}.
\newblock \bibinfo{booktitle}{\emph{The simulation heuristic}}.
\newblock


\bibitem[Khattar et~al\mbox{.}(2019)]%
        {khattar2019mvae}
\bibfield{author}{\bibinfo{person}{Dhruv Khattar}, \bibinfo{person}{Jaipal~Singh Goud}, \bibinfo{person}{Manish Gupta}, {and} \bibinfo{person}{Vasudeva Varma}.} \bibinfo{year}{2019}\natexlab{}.
\newblock \showarticletitle{Mvae: Multimodal variational autoencoder for fake news detection}. In \bibinfo{booktitle}{\emph{WWW}}. \bibinfo{pages}{2915--2921}.
\newblock


\bibitem[Kipf and Welling(2016)]%
        {kipf2016semi}
\bibfield{author}{\bibinfo{person}{Thomas~N Kipf} {and} \bibinfo{person}{Max Welling}.} \bibinfo{year}{2016}\natexlab{}.
\newblock \showarticletitle{Semi-supervised classification with graph convolutional networks}.
\newblock \bibinfo{journal}{\emph{arXiv preprint arXiv:1609.02907}} (\bibinfo{year}{2016}).
\newblock


\bibitem[Kulakova et~al\mbox{.}(2013)]%
        {kulakova2013processing}
\bibfield{author}{\bibinfo{person}{Eugenia Kulakova}, \bibinfo{person}{Markus Aichhorn}, \bibinfo{person}{Matthias Schurz}, \bibinfo{person}{Martin Kronbichler}, {and} \bibinfo{person}{Josef Perner}.} \bibinfo{year}{2013}\natexlab{}.
\newblock \showarticletitle{Processing counterfactual and hypothetical conditionals: An fMRI investigation}.
\newblock \bibinfo{journal}{\emph{NeuroImage}}  \bibinfo{volume}{72} (\bibinfo{year}{2013}), \bibinfo{pages}{265--271}.
\newblock


\bibitem[Lewis(1973)]%
        {lewis1973counterfactuals}
\bibfield{author}{\bibinfo{person}{David Lewis}.} \bibinfo{year}{1973}\natexlab{}.
\newblock \showarticletitle{Counterfactuals and comparative possibility}.
\newblock In \bibinfo{booktitle}{\emph{IFS: Conditionals, Belief, Decision, Chance and Time}}. \bibinfo{pages}{57--85}.
\newblock


\bibitem[Lewis(2013)]%
        {lewis2013counterfactuals}
\bibfield{author}{\bibinfo{person}{David Lewis}.} \bibinfo{year}{2013}\natexlab{}.
\newblock \bibinfo{booktitle}{\emph{Counterfactuals}}.
\newblock


\bibitem[Li et~al\mbox{.}(2021)]%
        {li2021braingnn}
\bibfield{author}{\bibinfo{person}{Xiaoxiao Li}, \bibinfo{person}{Yuan Zhou}, \bibinfo{person}{Nicha Dvornek}, \bibinfo{person}{Muhan Zhang}, \bibinfo{person}{Siyuan Gao}, \bibinfo{person}{Juntang Zhuang}, \bibinfo{person}{Dustin Scheinost}, \bibinfo{person}{Lawrence~H Staib}, \bibinfo{person}{Pamela Ventola}, {and} \bibinfo{person}{James~S Duncan}.} \bibinfo{year}{2021}\natexlab{}.
\newblock \showarticletitle{Braingnn: Interpretable brain graph neural network for fmri analysis}.
\newblock \bibinfo{journal}{\emph{Medical Image Analysis}}  \bibinfo{volume}{74} (\bibinfo{year}{2021}), \bibinfo{pages}{102233}.
\newblock


\bibitem[Li et~al\mbox{.}(2020)]%
        {li2020survey}
\bibfield{author}{\bibinfo{person}{Xiao-Hui Li}, \bibinfo{person}{Caleb~Chen Cao}, \bibinfo{person}{Yuhan Shi}, \bibinfo{person}{Wei Bai}, \bibinfo{person}{Han Gao}, \bibinfo{person}{Luyu Qiu}, \bibinfo{person}{Cong Wang}, \bibinfo{person}{Yuanyuan Gao}, \bibinfo{person}{Shenjia Zhang}, \bibinfo{person}{Xun Xue}, {et~al\mbox{.}}} \bibinfo{year}{2020}\natexlab{}.
\newblock \showarticletitle{A survey of data-driven and knowledge-aware explainable ai}.
\newblock \bibinfo{journal}{\emph{IEEE Transactions on Knowledge and Data Engineering}} \bibinfo{volume}{34}, \bibinfo{number}{1} (\bibinfo{year}{2020}), \bibinfo{pages}{29--49}.
\newblock


\bibitem[Lin et~al\mbox{.}(2023a)]%
        {lin2023multi}
\bibfield{author}{\bibinfo{person}{Fu Lin}, \bibinfo{person}{Haonan Gong}, \bibinfo{person}{Mingkang Li}, \bibinfo{person}{Zitong Wang}, \bibinfo{person}{Yue Zhang}, {and} \bibinfo{person}{Xuexiong Luo}.} \bibinfo{year}{2023}\natexlab{a}.
\newblock \showarticletitle{Multi-representations Space Separation based Graph-level Anomaly-aware Detection}. In \bibinfo{booktitle}{\emph{SSDBM}}. \bibinfo{pages}{1--11}.
\newblock


\bibitem[Lin et~al\mbox{.}(2023b)]%
        {lin2023discriminative}
\bibfield{author}{\bibinfo{person}{Fu Lin}, \bibinfo{person}{Xuexiong Luo}, \bibinfo{person}{Jia Wu}, \bibinfo{person}{Jian Yang}, \bibinfo{person}{Shan Xue}, \bibinfo{person}{Zitong Wang}, {and} \bibinfo{person}{Haonan Gong}.} \bibinfo{year}{2023}\natexlab{b}.
\newblock \showarticletitle{Discriminative Graph-Level Anomaly Detection via Dual-Students-Teacher Model}. In \bibinfo{booktitle}{\emph{ADMA}}. \bibinfo{pages}{261--276}.
\newblock


\bibitem[Liu et~al\mbox{.}(2021)]%
        {liu2021intention}
\bibfield{author}{\bibinfo{person}{Can Liu}, \bibinfo{person}{Li Sun}, \bibinfo{person}{Xiang Ao}, \bibinfo{person}{Jinghua Feng}, \bibinfo{person}{Qing He}, {and} \bibinfo{person}{Hao Yang}.} \bibinfo{year}{2021}\natexlab{}.
\newblock \showarticletitle{Intention-aware heterogeneous graph attention networks for fraud transactions detection}. In \bibinfo{booktitle}{\emph{KDD}}. \bibinfo{pages}{3280--3288}.
\newblock


\bibitem[Liu et~al\mbox{.}(2008)]%
        {liu2008isolation}
\bibfield{author}{\bibinfo{person}{Fei~Tony Liu}, \bibinfo{person}{Kai~Ming Ting}, {and} \bibinfo{person}{Zhi-Hua Zhou}.} \bibinfo{year}{2008}\natexlab{}.
\newblock \showarticletitle{Isolation forest}. In \bibinfo{booktitle}{\emph{ICDM}}. \bibinfo{pages}{413--422}.
\newblock


\bibitem[Liu et~al\mbox{.}(2023)]%
        {liu2023good}
\bibfield{author}{\bibinfo{person}{Yixin Liu}, \bibinfo{person}{Kaize Ding}, \bibinfo{person}{Huan Liu}, {and} \bibinfo{person}{Shirui Pan}.} \bibinfo{year}{2023}\natexlab{}.
\newblock \showarticletitle{GOOD-D: On Unsupervised Graph Out-Of-Distribution Detection}. In \bibinfo{booktitle}{\emph{WSDM}}. \bibinfo{pages}{339--347}.
\newblock


\bibitem[Liu et~al\mbox{.}(2024)]%
        {liu2024towards}
\bibfield{author}{\bibinfo{person}{Yixin Liu}, \bibinfo{person}{Kaize Ding}, \bibinfo{person}{Qinghua Lu}, \bibinfo{person}{Fuyi Li}, \bibinfo{person}{Leo~Yu Zhang}, {and} \bibinfo{person}{Shirui Pan}.} \bibinfo{year}{2024}\natexlab{}.
\newblock \showarticletitle{Towards self-interpretable graph-level anomaly detection}.
\newblock \bibinfo{journal}{\emph{NeurIPS}}  \bibinfo{volume}{36} (\bibinfo{year}{2024}).
\newblock


\bibitem[Luo et~al\mbox{.}(2024)]%
        {luo2024graph}
\bibfield{author}{\bibinfo{person}{Xuexiong Luo}, \bibinfo{person}{Jia Wu}, \bibinfo{person}{Jian Yang}, \bibinfo{person}{Shan Xue}, \bibinfo{person}{Amin Beheshti}, \bibinfo{person}{Quan~Z Sheng}, \bibinfo{person}{David McAlpine}, \bibinfo{person}{Paul Sowman}, \bibinfo{person}{Alexis Giral}, {and} \bibinfo{person}{Philip~S Yu}.} \bibinfo{year}{2024}\natexlab{}.
\newblock \showarticletitle{Graph Neural Networks for Brain Graph Learning: A Survey}.
\newblock \bibinfo{journal}{\emph{arXiv preprint arXiv:2406.02594}} (\bibinfo{year}{2024}).
\newblock


\bibitem[Luo et~al\mbox{.}(2022)]%
        {luo2022deep}
\bibfield{author}{\bibinfo{person}{Xuexiong Luo}, \bibinfo{person}{Jia Wu}, \bibinfo{person}{Jian Yang}, \bibinfo{person}{Shan Xue}, \bibinfo{person}{Hao Peng}, \bibinfo{person}{Chuan Zhou}, \bibinfo{person}{Hongyang Chen}, \bibinfo{person}{Zhao Li}, {and} \bibinfo{person}{Quan~Z Sheng}.} \bibinfo{year}{2022}\natexlab{}.
\newblock \showarticletitle{Deep graph level anomaly detection with contrastive learning}.
\newblock \bibinfo{journal}{\emph{Scientific Reports}} \bibinfo{volume}{12}, \bibinfo{number}{1} (\bibinfo{year}{2022}), \bibinfo{pages}{19867}.
\newblock


\bibitem[Ma et~al\mbox{.}(2022)]%
        {ma2022deep}
\bibfield{author}{\bibinfo{person}{Rongrong Ma}, \bibinfo{person}{Guansong Pang}, \bibinfo{person}{Ling Chen}, {and} \bibinfo{person}{Anton van~den Hengel}.} \bibinfo{year}{2022}\natexlab{}.
\newblock \showarticletitle{Deep graph-level anomaly detection by glocal knowledge distillation}. In \bibinfo{booktitle}{\emph{WSDM}}. \bibinfo{pages}{704--714}.
\newblock


\bibitem[McGill and Klein(1993)]%
        {mcgill1993contrastive}
\bibfield{author}{\bibinfo{person}{Ann~L McGill} {and} \bibinfo{person}{Jill~G Klein}.} \bibinfo{year}{1993}\natexlab{}.
\newblock \showarticletitle{Contrastive and counterfactual reasoning in causal judgment.}
\newblock \bibinfo{journal}{\emph{Journal of Personality and Social Psychology}} \bibinfo{volume}{64}, \bibinfo{number}{6} (\bibinfo{year}{1993}), \bibinfo{pages}{897}.
\newblock


\bibitem[Morris et~al\mbox{.}(2020)]%
        {morris2020tudataset}
\bibfield{author}{\bibinfo{person}{Christopher Morris}, \bibinfo{person}{Nils~M Kriege}, \bibinfo{person}{Franka Bause}, \bibinfo{person}{Kristian Kersting}, \bibinfo{person}{Petra Mutzel}, {and} \bibinfo{person}{Marion Neumann}.} \bibinfo{year}{2020}\natexlab{}.
\newblock \showarticletitle{Tudataset: A collection of benchmark datasets for learning with graphs}.
\newblock \bibinfo{journal}{\emph{arXiv preprint arXiv:2007.08663}} (\bibinfo{year}{2020}).
\newblock


\bibitem[Pang et~al\mbox{.}(2019)]%
        {pang2019deep}
\bibfield{author}{\bibinfo{person}{Guansong Pang}, \bibinfo{person}{Chunhua Shen}, {and} \bibinfo{person}{Anton Van Den~Hengel}.} \bibinfo{year}{2019}\natexlab{}.
\newblock \showarticletitle{Deep anomaly detection with deviation networks}. In \bibinfo{booktitle}{\emph{KDD}}. \bibinfo{pages}{353--362}.
\newblock


\bibitem[Pearl and Mackenzie(2018)]%
        {pearl2018book}
\bibfield{author}{\bibinfo{person}{Judea Pearl} {and} \bibinfo{person}{Dana Mackenzie}.} \bibinfo{year}{2018}\natexlab{}.
\newblock \bibinfo{booktitle}{\emph{The book of why: the new science of cause and effect}}.
\newblock


\bibitem[Qiu et~al\mbox{.}(2022)]%
        {qiu2022raising}
\bibfield{author}{\bibinfo{person}{Chen Qiu}, \bibinfo{person}{Marius Kloft}, \bibinfo{person}{Stephan Mandt}, {and} \bibinfo{person}{Maja Rudolph}.} \bibinfo{year}{2022}\natexlab{}.
\newblock \showarticletitle{Raising the bar in graph-level anomaly detection}.
\newblock \bibinfo{journal}{\emph{arXiv preprint arXiv:2205.13845}} (\bibinfo{year}{2022}).
\newblock


\bibitem[Sch{\"o}lkopf et~al\mbox{.}(1999)]%
        {scholkopf1999support}
\bibfield{author}{\bibinfo{person}{Bernhard Sch{\"o}lkopf}, \bibinfo{person}{Robert~C Williamson}, \bibinfo{person}{Alex Smola}, \bibinfo{person}{John Shawe-Taylor}, {and} \bibinfo{person}{John Platt}.} \bibinfo{year}{1999}\natexlab{}.
\newblock \showarticletitle{Support vector method for novelty detection}. In \bibinfo{booktitle}{\emph{NeurIPS}}, Vol.~\bibinfo{volume}{12}.
\newblock


\bibitem[Shu et~al\mbox{.}(2017)]%
        {shu2017fake}
\bibfield{author}{\bibinfo{person}{Kai Shu}, \bibinfo{person}{Amy Sliva}, \bibinfo{person}{Suhang Wang}, \bibinfo{person}{Jiliang Tang}, {and} \bibinfo{person}{Huan Liu}.} \bibinfo{year}{2017}\natexlab{}.
\newblock \showarticletitle{Fake news detection on social media: A data mining perspective}. In \bibinfo{booktitle}{\emph{KDD}}, Vol.~\bibinfo{volume}{19}. \bibinfo{pages}{22--36}.
\newblock


\bibitem[Stepin et~al\mbox{.}(2021)]%
        {stepin2021survey}
\bibfield{author}{\bibinfo{person}{Ilia Stepin}, \bibinfo{person}{Jose~M Alonso}, \bibinfo{person}{Alejandro Catala}, {and} \bibinfo{person}{Mart{\'\i}n Pereira-Fari{\~n}a}.} \bibinfo{year}{2021}\natexlab{}.
\newblock \showarticletitle{A survey of contrastive and counterfactual explanation generation methods for explainable artificial intelligence}.
\newblock \bibinfo{journal}{\emph{IEEE Access}}  \bibinfo{volume}{9} (\bibinfo{year}{2021}), \bibinfo{pages}{11974--12001}.
\newblock


\bibitem[Tang et~al\mbox{.}(2022)]%
        {tang2022rethinking}
\bibfield{author}{\bibinfo{person}{Jianheng Tang}, \bibinfo{person}{Jiajin Li}, \bibinfo{person}{Ziqi Gao}, {and} \bibinfo{person}{Jia Li}.} \bibinfo{year}{2022}\natexlab{}.
\newblock \showarticletitle{Rethinking graph neural networks for anomaly detection}. In \bibinfo{booktitle}{\emph{PMLR}}. \bibinfo{pages}{21076--21089}.
\newblock


\bibitem[Veli{\v{c}}kovi{\'c} et~al\mbox{.}(2017)]%
        {velivckovic2017graph}
\bibfield{author}{\bibinfo{person}{Petar Veli{\v{c}}kovi{\'c}}, \bibinfo{person}{Guillem Cucurull}, \bibinfo{person}{Arantxa Casanova}, \bibinfo{person}{Adriana Romero}, \bibinfo{person}{Pietro Lio}, {and} \bibinfo{person}{Yoshua Bengio}.} \bibinfo{year}{2017}\natexlab{}.
\newblock \showarticletitle{Graph attention networks}.
\newblock \bibinfo{journal}{\emph{arXiv preprint arXiv:1710.10903}} (\bibinfo{year}{2017}).
\newblock


\bibitem[Verma and Zhang(2017)]%
        {verma2017hunt}
\bibfield{author}{\bibinfo{person}{Saurabh Verma} {and} \bibinfo{person}{Zhi-Li Zhang}.} \bibinfo{year}{2017}\natexlab{}.
\newblock \showarticletitle{Hunt for the unique, stable, sparse and fast feature learning on graphs}. In \bibinfo{booktitle}{\emph{NeurIPS}}, Vol.~\bibinfo{volume}{30}.
\newblock


\bibitem[Wang et~al\mbox{.}(2019)]%
        {wang2019semi}
\bibfield{author}{\bibinfo{person}{Daixin Wang}, \bibinfo{person}{Jianbin Lin}, \bibinfo{person}{Peng Cui}, \bibinfo{person}{Quanhui Jia}, \bibinfo{person}{Zhen Wang}, \bibinfo{person}{Yanming Fang}, \bibinfo{person}{Quan Yu}, \bibinfo{person}{Jun Zhou}, \bibinfo{person}{Shuang Yang}, {and} \bibinfo{person}{Yuan Qi}.} \bibinfo{year}{2019}\natexlab{}.
\newblock \showarticletitle{A semi-supervised graph attentive network for financial fraud detection}. In \bibinfo{booktitle}{\emph{ICDM}}. \bibinfo{pages}{598--607}.
\newblock


\bibitem[Wu et~al\mbox{.}(2020)]%
        {wu2020comprehensive}
\bibfield{author}{\bibinfo{person}{Zonghan Wu}, \bibinfo{person}{Shirui Pan}, \bibinfo{person}{Fengwen Chen}, \bibinfo{person}{Guodong Long}, \bibinfo{person}{Chengqi Zhang}, {and} \bibinfo{person}{S~Yu Philip}.} \bibinfo{year}{2020}\natexlab{}.
\newblock \showarticletitle{A comprehensive survey on graph neural networks}.
\newblock \bibinfo{journal}{\emph{IEEE Transactions on Neural Networks and Learning Systems}} \bibinfo{volume}{32}, \bibinfo{number}{1} (\bibinfo{year}{2020}), \bibinfo{pages}{4--24}.
\newblock


\bibitem[Yang et~al\mbox{.}(2023)]%
        {yang2023generating}
\bibfield{author}{\bibinfo{person}{Haoran Yang}, \bibinfo{person}{Hongxu Chen}, \bibinfo{person}{Sixiao Zhang}, \bibinfo{person}{Xiangguo Sun}, \bibinfo{person}{Qian Li}, \bibinfo{person}{Xiangyu Zhao}, {and} \bibinfo{person}{Guandong Xu}.} \bibinfo{year}{2023}\natexlab{}.
\newblock \showarticletitle{Generating counterfactual hard negative samples for graph contrastive learning}. In \bibinfo{booktitle}{\emph{WWW}}. \bibinfo{pages}{621--629}.
\newblock


\bibitem[Ying et~al\mbox{.}(2018)]%
        {ying2018hierarchical}
\bibfield{author}{\bibinfo{person}{Zhitao Ying}, \bibinfo{person}{Jiaxuan You}, \bibinfo{person}{Christopher Morris}, \bibinfo{person}{Xiang Ren}, \bibinfo{person}{Will Hamilton}, {and} \bibinfo{person}{Jure Leskovec}.} \bibinfo{year}{2018}\natexlab{}.
\newblock \showarticletitle{Hierarchical graph representation learning with differentiable pooling}.
\newblock \bibinfo{journal}{\emph{NeurIPS}}  \bibinfo{volume}{31} (\bibinfo{year}{2018}).
\newblock


\bibitem[You et~al\mbox{.}(2020)]%
        {you2020graph}
\bibfield{author}{\bibinfo{person}{Yuning You}, \bibinfo{person}{Tianlong Chen}, \bibinfo{person}{Yongduo Sui}, \bibinfo{person}{Ting Chen}, \bibinfo{person}{Zhangyang Wang}, {and} \bibinfo{person}{Yang Shen}.} \bibinfo{year}{2020}\natexlab{}.
\newblock \showarticletitle{Graph contrastive learning with augmentations}.
\newblock \bibinfo{journal}{\emph{NeurIPS}}  \bibinfo{volume}{33} (\bibinfo{year}{2020}), \bibinfo{pages}{5812--5823}.
\newblock


\bibitem[Zhao and Akoglu(2021)]%
        {zhao2021glod-issues}
\bibfield{author}{\bibinfo{person}{Lingxiao Zhao} {and} \bibinfo{person}{Leman Akoglu}.} \bibinfo{year}{2021}\natexlab{}.
\newblock \showarticletitle{On using classification datasets to evaluate graph outlier detection: Peculiar observations and new insights}.
\newblock \bibinfo{journal}{\emph{Big Data}} (\bibinfo{year}{2021}).
\newblock


\bibitem[Zhao and Akoglu(2023)]%
        {zhao2023using}
\bibfield{author}{\bibinfo{person}{Lingxiao Zhao} {and} \bibinfo{person}{Leman Akoglu}.} \bibinfo{year}{2023}\natexlab{}.
\newblock \showarticletitle{On using classification datasets to evaluate graph outlier detection: Peculiar observations and new insights}.
\newblock \bibinfo{journal}{\emph{Big Data}} \bibinfo{volume}{11}, \bibinfo{number}{3} (\bibinfo{year}{2023}), \bibinfo{pages}{151--180}.
\newblock


\end{thebibliography}

\end{document}